\documentclass{article}

\PassOptionsToPackage{numbers,sort&compress}{natbib}
\usepackage[preprint]{neurips_2026}

\usepackage[utf8]{inputenc} 
\usepackage[T1]{fontenc}    
\usepackage{hyperref}       
\usepackage{url}            
\usepackage{booktabs}       
\usepackage{amsfonts}       
\usepackage{nicefrac}       
\usepackage{microtype}      
\usepackage{xcolor}         
\usepackage{listings}
\usepackage{booktabs}
\usepackage{graphicx}
\usepackage{multirow}
\usepackage{amsmath}
\usepackage{pifont}
\usepackage{pgfplots}
\usepackage{subcaption}
\usepackage{geometry}
\usepackage{makecell}
\usepackage{caption}
\usepackage{algorithm}
\usepackage{algpseudocode}
\usepackage{enumitem}
\pgfplotsset{compat=1.18}

\lstdefinelanguage{prompt}{
  basicstyle=\ttfamily\small,
  breaklines=true,
  breakindent=0pt,
  breakautoindent=false,
  frame=single,
  backgroundcolor=\color{pink!10},
  columns=fullflexible,
  showstringspaces=false,
  moredelim=**[is][\color{blue}]{<<}{>>}
}

\title{MM-DyGraph: A Dataset, Benchmark, and Model for Multimodal Dynamic Graphs}

\author{
  \normalfont
  \begin{tabular}{@{}cccc@{}}
    \textbf{Zongyuan Wu}\textsuperscript{1}
      \thanks{These authors contributed equally to this research.} &
    \textbf{Pengjie Wang}\textsuperscript{1}\footnotemark[1] &
    \textbf{Lingshan Chen}\textsuperscript{2}\footnotemark[1] &
    \textbf{Tianhang Wan}\textsuperscript{1}
    \\[0.4em]
    \textbf{Yuan Meng}\textsuperscript{1}
      \thanks{Corresponding authors.} &
    \textbf{Xin Wang}\textsuperscript{1}\footnotemark[2] &
    \textbf{Ling Feng}\textsuperscript{1}\footnotemark[2] &
    \textbf{Wenwu Zhu}\textsuperscript{1}\footnotemark[2]
  \end{tabular}\\[0.5em]
  \textsuperscript{1}DCST \quad
  \textsuperscript{2}Zhili College\\
  Tsinghua University, Beijing, China\\[0.3em]
  \texttt{\{wuzongyu24,wpj24,cls24,wanth21\}@mails.tsinghua.edu.cn}\\
  \texttt{yuanmeng@mail.tsinghua.edu.cn}\\
  \texttt{\{xin\_wang,fengling,wwzhu\}@tsinghua.edu.cn}
}

\begin{document}

\maketitle

\begin{abstract}
    Multimodal dynamic graphs are pervasive in real-world applications, where nodes and edges are accompanied by multimodal signals, including text and images. Despite their practical importance, benchmarks and methods that explicitly study multimodal learning in dynamic graphs remain scarce, hindering the full understanding and effective utilization of multimodal information for downstream tasks.
    To address this gap, we introduce \textbf{MM-DyGraph}\footnote{Dataset is available on \url{https://huggingface.co/datasets/LeslieWu999/MM-DyGraph}} \footnote{Code is available on \url{https://github.com/LeslieWu999/mm-dygraph}}, to the best of our knowledge, the first benchmark for multimodal dynamic graph learning, consisting of diverse real-world datasets with fine-grained temporal graph evolution and rich multimodal attributes. 
    \textbf{MM-DyGraph} provides three evaluation tasks to comprehensively assess model performance in dynamic settings. We benchmark representative state-of-the-art dynamic graph methods and reveal that naive feature concatenation yields \textit{target-level inconsistencies}: multimodal inputs benefit certain prediction targets while harming others.
    This observation indicates that the modality informativeness depends on the specific prediction target over time. Building on these observations, we propose a \underline{\textbf{T}}arget-aware \underline{\textbf{S}}patial-\underline{\textbf{T}}emporal \underline{\textbf{M}}ultimodal \underline{\textbf{F}}usion (\textbf{TST-MF}) framework, which jointly conditions multimodal fusion on the prediction target and spatial-temporal relations, enabling context-adaptive cross-modal interplay for target-specific representations. Extensive experiments over diverse datasets and tasks show that \textbf{TST-MF} delivers robust and competitive performance, achieving clear gains in multimodal settings.
\end{abstract}

\section{Introduction}
Dynamic graphs provide a natural abstraction for evolving relational systems, where nodes and edges evolve continuously over time.
Such data are ubiquitous in high-impact applications including social platforms~\cite{DBLP:conf/iclr/TrivediFBZ19, rossi2020temporalgraphnetworksdeep, xu2020inductiverepresentationlearningtemporal, Kumar_2019}, recommendation and advertising systems~\cite{10.1145/3701988,9714053, 10.1145/3769103, rajabi2021clickthroughratepredictionusing,10.1145/3593314}, and finance~\cite{kulkarni2025dynbergdynamicbertbasedgraph, duan2024catgnnenhancingcreditcard, kim2024temporalgraphnetworksgraph}, motivating a rich family of temporal graph neural networks to capture graph dynamics and spatiotemporal dependencies~\cite{NEURIPS2022_2857242c,zhang2024spectralinvariantlearningdynamic,wang2022inductiverepresentationlearningtemporal,wang2021tcltransformerbaseddynamicgraph, cong2023reallyneedcomplicatedmodel, yu2023betterdynamicgraphlearning, Zhang_Zhang_Wang_Qin_Qin_Zhu_2023}. 
However, real-world dynamic graphs are rarely structure-only. 
Nodes and edges are often accompanied by multimodal content signals, such as product titles and images in e-commerce and advertising~\cite{sheng2024enhancingtaobaodisplayadvertising}, short transaction descriptions and document-like evidence in payment ecosystems~\cite{10.1007/s10032-022-00399-3}, and multimodal posts in social platforms that shape diffusion and engagement~\cite{10.1145/3447270}. 
These multimodal attributes provide complementary semantics that play a vital role in improving downstream task performance, particularly when structural connectivity is sparse, noisy, or rapidly changing~\cite{9428193,wei2024gitagraphvisualtextual}. For example, a product's appearance and textual description may contribute differently to predicting a user's next purchase as their interests and interaction history change. Such applications call for the joint modeling of multimodal content and dynamic graph structure to capture how semantic information informs predictions in evolving relational contexts.

Despite this practical importance, benchmarks and methods that systematically study multimodal learning on dynamic graphs remain scarce. 
Existing benchmarks largely focus on structural and textual attributes~\cite{jin2024largelanguagemodelsgraphs, NEURIPS2023_37d00f56, zhang2024dtgbcomprehensivebenchmarkdynamic}, leaving visual signals underexplored in spatiotemporal evolution.
To the best of our knowledge, no existing benchmark incorporates multimodal information for dynamic graphs, making it infeasible to develop and evaluate corresponding methods.

To address this gap, we introduce \textbf{MM-DyGraph}, a benchmark for multimodal dynamic graph learning. It consists of diverse real-world datasets with fine-grained temporal graph evolution and rich multimodal attributes and provides three evaluation tasks, \emph{link prediction}, \emph{edge classification}, and \emph{node retrieval}, to comprehensively assess model performance in dynamic settings. Unlike prior benchmarks that largely restrict node and edge attributes to a single modality, \textbf{MM-DyGraph} enables systematic evaluation when multiple modalities (e.g., text and images) are simultaneously available, thereby facilitating fair comparisons and encouraging progress toward realistic multimodal dynamic graph learning.

With \textbf{MM-DyGraph}, we benchmark representative state-of-the-art dynamic graph learning methods and observe that naively concatenating multimodal features does not reliably improve performance.
Furthermore, we find \textit{target-level inconsistencies}: multimodal inputs may benefit certain prediction targets (e.g., specific nodes or edges) while harming others, indicating that modality utility is target-dependent and intertwined with spatial neighborhood and temporal history. Motivated by recent advances in context-adaptive multimodal modeling~\cite{guo2026phaselora,he2026enhancethenbalancemodalitycollaborationrobust,wu2025admnlayerwiseadaptivemultimodal,park2025resilientsensorfusionadverse}, we hypothesize that a key underlying reason is that unified fusion strategies treat multimodal signals as static, equally informative features, which can be insufficient for multimodal dynamic graphs, where the same modality may be informative for some targets but distracting for others under temporal evolution.

To address this problem, we propose \textbf{TST-MF}, a \underline{\textbf{T}}arget-aware \underline{\textbf{S}}patial-\underline{\textbf{T}}emporal \underline{\textbf{M}}ultimodal \underline{\textbf{F}}usion framework for multimodal dynamic graphs. Rather than applying a fixed fusion rule, \textbf{TST-MF} performs context-adaptive cross-modal interplay by jointly conditioning fusion on (i) the prediction target, (ii) the target's spatial relations, and (iii) the target's temporal context. This design enables the model to emphasize task-relevant multimodal cues for each target while suppressing redundant or noisy signals, yielding more robust representations under spatiotemporal evolution. Extensive experiments across datasets and tasks demonstrate that \textbf{TST-MF} achieves clear gains over strong baselines, validating the effectiveness and generality of target- and context-conditioned fusion for multimodal dynamic graphs.

In summary, we summarize our contributions as follows:
\begin{enumerate}[leftmargin=*, nosep]
    \item \textbf{First Multimodal Dynamic Graph Benchmark:} To the best of our knowledge, we propose \textbf{MM-DyGraph}, the first benchmark for multimodal dynamic graph learning with diverse real-world datasets, rich multimodal attributes, and three downstream evaluation tasks. This benchmark will facilitate the development of multimodal dynamic graph learning methods and advance performance on downstream tasks.
    \item \textbf{Key Observations:} We systematically benchmark representative state-of-the-art methods and reveal that naive multimodal fusion yields highly inconsistent effects across different targets. The same modality can be informative for some nodes but detrimental for others. This challenges the assumption that uniform fusion universally benefits all targets.
    \item \textbf{Effective Model:} We propose \textbf{TST-MF}, a novel target-aware spatiotemporal multimodal fusion framework that strategically conditions cross-modal interplay on the prediction target and its spatiotemporal context. By seamlessly integrating multimodal information, \textbf{TST-MF} achieves substantial improvements across diverse datasets and tasks, offering an effective and principled solution for multimodal dynamic graph learning.
\end{enumerate}

\section{Related Work}
\subsection{Benchmarks for Dynamic Graph Representation Learning}
Benchmarking has played a critical role in advancing dynamic graph learning by providing standardized datasets, evaluation protocols, and reproducible pipelines. Early efforts highlighted that evaluation for temporal link prediction is often confounded by edge recurrence and unrealistic data splits, motivating more rigorous protocols tailored to dynamic graphs~\cite{dgb_neurips_D&B_2022,poursafaei2022betterevaluationdynamiclink,10.1145/3637528.3672001}. More recently, the Temporal Graph Benchmark (TGB)~\cite{huang2023temporalgraphbenchmarkmachine} established a diverse suite of large-scale temporal interaction datasets spanning multiple domains, together with realistic evaluation protocols for node- and edge-level prediction tasks and an end-to-end benchmarking pipeline. To further support multi-relational temporal settings, TGB 2.0~\cite{gastinger2024tgb20benchmarklearning} extended benchmarking to temporal knowledge graphs and heterogeneous temporal graphs, emphasizing scalability and reproducibility in large-scale multi-relational link prediction. In parallel, benchmarks have begun to incorporate richer attributes beyond structure. DTGB~\cite{zhang2024dtgbcomprehensivebenchmarkdynamic} focuses on dynamic text-attributed graphs and provides standardized evaluation under multiple real-world use cases, enabling systematic study of temporal graph methods with evolving textual attributes. While these benchmarks significantly improve the evaluation landscape for dynamic graphs, they predominantly target unimodal attributes. 
This limits the evaluation and model design biased toward unimodal cues, which can lead to suboptimal downstream performance in realistic settings where multimodal signals, especially visual information, are essential under temporal evolution.

\subsection{Multimodal Graph Learning}
Multimodal graph learning studies how to combine heterogeneous modalities (i.e., text, images, and other side information) with graph structure to improve downstream tasks \cite{wang2025multimodalgraphlargelanguage}. Existing research in this area has primarily focused on static graphs, where multimodal node and edge attributes are fused through feature-level concatenation, modality-specific encoders followed by graph message passing, or cross-modal alignment modules that enforce semantic consistency across modalities~\cite{peng2024learningmultimodalgraphssurvey}. A major line of work focuses on multimodal attributed graphs, where nodes and edges carry multimodal attributes that are encoded by modality-specific encoders and fused via concatenation, attention, or gating before or within graph propagation \cite{yan2025graphmeetsmultimodalbenchmarking,zhu2025mosaicmodalities,MMGCN,LI2024119815,Cai_Wang_Li_Zhang_Zhu_2024}. Another closely related direction enriches knowledge graphs with multimodal entity information, incorporating visual and textual signals into knowledge graph embeddings or scoring functions to improve knowledge graph completion and link prediction \cite{wang2023tiva,WANG202278,zhu2025mosaicmodalities,10.24963/ijcai.2024/659,TRAN2025113541}. Despite encouraging progress, most existing methods and benchmarks are designed for static graphs, leaving multimodal fusion under spatiotemporal evolution in dynamic graphs less explored.

\section{MM-DyGraph}

\begin{figure}
    \centering
    \includegraphics[width=\linewidth]{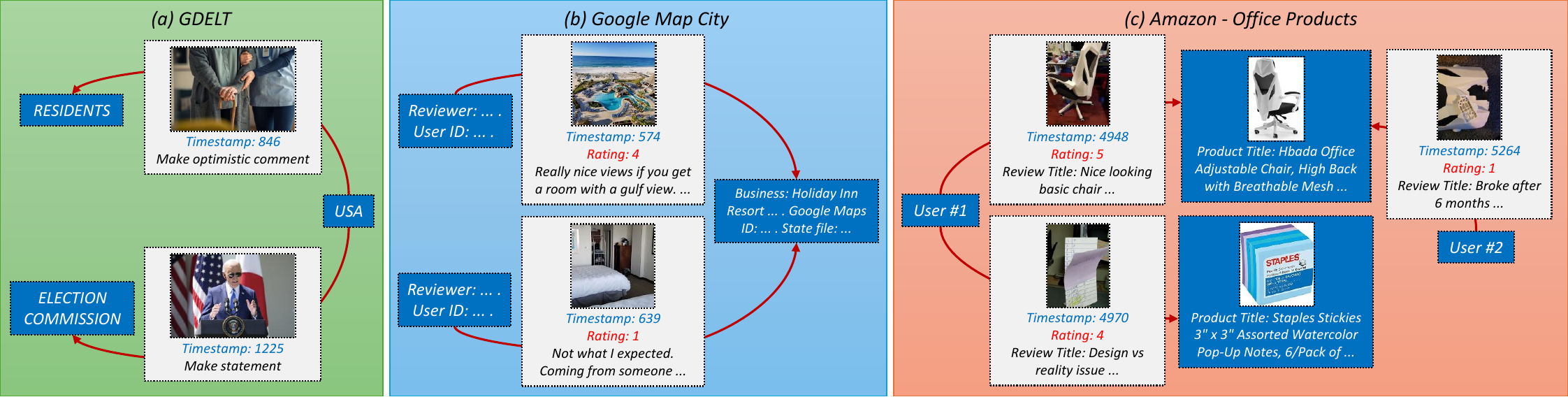}
    \caption{Visualization of \textbf{MM-DyGraph}, in which multimodal information can exist on \textit{both} nodes and edges.}
    \label{fig:benchmark_visualization}
\end{figure}

In this section, we present the proposed real-world datasets in \textbf{MM-DyGraph} (a visualization is presented in Figure \ref{fig:benchmark_visualization}). The remaining sections are structured as follows: the data sources are described in Section \ref{sec:drdc}, the graph construction process is mentioned in Section \ref{sec:ppgc}, the evaluation tasks are given in Section \ref{sec:tasks}, and the data statistics are introduced in Section \ref{sec:statistics}.

\subsection{Data Sources}
\label{sec:drdc}

\textbf{GDELT} \cite{Leetaru2013GDELT} focuses on sociopolitical event extraction. It is modeled as a heterogeneous, non-bipartite and temporal knowledge graph, where nodes represent named entities, and edges denote directed sociopolitical interactions encoded via the CAMEO (Conflict and Mediation Event Observations) taxonomy. Visual information is introduced and obtained by extracting figures from the website through Beautiful Soup.

\noindent
\textbf{Google Map City (GMCT)} \cite{li-etal-2022-uctopic, 10.1145/3539618.3592036} focuses on point-of-interest recommendation, featuring spatial-temporal data mining and urban computing. It is modeled as a bipartite and highly sparse graph, containing multimodal features ranging from geo-coordinates to review text and other multimodal data. Visual information is introduced and obtained by downloading figures through links in the metadata. It provides a sparse interaction setting with multimodal content attached to edges and limited user-side attributes, complementing datasets with richer node-level features.

\noindent
\textbf{Amazon} \cite{hou2024bridging} focuses on E-commerce recommendation. It is also modeled as a bipartite and highly sparse graph, aggregating massive logs of user-item interactions metadata and user-written contents, which allow the modeling of sequential dynamics. Visual information is introduced and obtained by downloading figures through links provided by the metadata.

\noindent
\textbf{Yelp} \cite{YelpOpenDataset} is a dataset focusing on social network analysis and local business recommendation, whose graphs are heterogeneous and bipartite. It offers a comprehensive view of local business interactions. Visual information is originally provided by the dataset.

\subsection{Dataset Construction}
\label{sec:ppgc}

To ensure the quality of the graph data and the reliability of cross-modal associations, we implement a comprehensive post-processing pipeline following the initial data acquisition. This pipeline involves dataset-specific pruning to increase multimodal information density, and Vision-Language Model (VLM)-based quality assessment to filter out noisy or irrelevant figures.

The dataset-specific pruning procedure aims to accumulate sufficient cross-modal information for multimodal learning, enabling controlled evaluation without letting missing-media handling become the primary confounder. For Amazon, we exclude user nodes that have not contributed any image-containing reviews. For GMCT, we implement merchant-level pruning by removing business nodes whose edges lack visual documentation, whereas on Yelp, the pruning targets business nodes without attached images. GDELT remains unchanged.

After pruning, the datasets acquire a higher density of visual information, while also obtaining a size suitable for benchmarking. To support research under different modality-availability conditions, the published dataset includes both the pruned benchmark version and a complete version that retains all interactions within the same source scope prior to pruning. Details are given in Appendix \ref{app:statistic}.

The VLM-based quality assessment focuses on cross-modal relevance, where we utilize Qwen3-VL-8B-Instruct \cite{bai2025qwen3vltechnicalreport}. Given an image-text pair, the VLM is prompted to generate a brief reasoning trace before assigning a relevance score for the pair, ranging from 0 to 10. The relevance score captures the broad semantic correspondence between the image and text, offering an accurate criterion to assess the cross-modal correlation. Manual supervision has validated the reliability of this VLM assessment (some case studies are provided in Appendix \ref{app:casestudy}). We filter out images with a relevance score lower than 6. If there still remain multiple images for a certain text after filtering, the one with the highest relevance score is retained. The prompt used in the assessment is presented in Appendix \ref{app: prompt}.

To facilitate temporal sequence modeling, raw timestamps are transformed into discrete time steps via a temporal bucketing scheme, formed as $\tau = \left\lfloor \frac{t_{raw} - t_{min}}{\Delta t} \right\rfloor$, where $\Delta t$ is the discretization granularity that adapted based on the dataset characteristics to capture temporal dynamics inherent in different datasets. $\Delta t = 15\text{min}$ is applied for datasets characterizing rapid state transitions, namely Google Map City and GDELT, while a coarser interval of $\Delta t = 24\text{h}$ is utilized for datasets driven with daily periodicity and macro-level behavioral trends, namely Amazon and Yelp.

\subsection{Tasks}
\label{sec:tasks}

\textbf{Link Prediction} evaluates models' performance on predicting whether there are links between provided node pairs. For each batch, real source-destination pairs are treated as positive examples, while negative examples are generated by a negative edge sampler. The model computes temporal representations for both positive and negative pairs and predicts their link probabilities. Following DTGB \cite{zhang2024dtgbcomprehensivebenchmarkdynamic}, interactions are ordered chronologically and divided into training, validation, and test sets with a 70\%/15\%/15\% split and adopt its transductive and inductive evaluation settings. Transductive evaluation measures performance on future interactions, while inductive evaluation focuses on interactions involving nodes absent from the training graph. Performance is measured by Average Precision (AP) and Area Under the Receiver Operating Characteristic Curve (AUC-ROC).

\textbf{Edge Classification} evaluates models' performance on predicting the categorical label of an observed temporal edge. Given a node pair, the interaction timestamp, and the edge identifier, the model computes temporal node-pair representations and predicts an edge-label distribution with the classification head. Performance is evaluated using Precision, Recall, and F1.

\textbf{Node Retrieval} is a ranking evaluation built on the trained link prediction model. For each real source-destination interaction, the true destination node is treated as the ground truth, with 100 sampled ones as negative candidates. The model scores each source-destination pair with the same link probability predictor used for link prediction. Results on transductive and inductive settings are reported. Hits@$k$ is utilized as the evaluation metric.

\subsection{Statistics}
\label{sec:statistics}
We summarize the statistics of the previous and our proposed datasets in Appendix~\ref{app:statistic}. In our dataset, multimodal information is provided on both nodes and edges, where \textbf{GDELT} and \textbf{Google Map City} contain images only on edges, \textbf{Yelp} provides images only on nodes, and \textbf{Amazon} includes images on both nodes and edges. Other statistical analyses are presented in Appendix~\ref{app:dataset}.

\section{Empirical Analysis}

In this section, we first present the evaluation framework of \textbf{MM-DyGraph}, then conduct experiments to analyze the following three research questions:

\textbf{(RQ1)} Can existing dynamic graph methods work for multimodal situations?

\textbf{(RQ2)} Do additional modalities and additional information guarantee better performance?

\textbf{(RQ3)} Do different prediction targets benefit equally from different modalities?

\subsection{Evaluation Settings}
\label{sec:evaluationsetting}

\textbf{Embedding Method.} Textual and visual information is encoded by a CLIP \cite{radford2021learningtransferablevisualmodels} model \footnote{openai/clip-vit-base-patch16}, which maps each piece of text and each image to a 512-dimensional vector. This model possesses excellent modality alignment capabilities that facilitate the subsequent graph learning process. Due to CLIP's intrinsic input text length limit, a chunking-and-averaging strategy is implemented to obtain the overall text embedding. We first encode the modality information and then provide the raw embeddings to the graph model. To enhance modal alignment, we normalize the raw embeddings for each modality separately.

\textbf{Features.} There are three feature configurations to evaluate the impact of multimodal integration: (i) random embedding baseline, (ii) unimodal (textual) embedding, and (iii) multimodal (visual-textual) representation. For the visual-textual configuration, we concatenate the text embedding and the image embedding to form a fused 1024-dimensional embedding. To ensure dimensional consistency across diverse input types, for the unimodal configuration, we duplicate the text embedding twice to match the fused feature space. In cases where the visual modality is not available, the information gap is mitigated by replacing the missing components with the corresponding textual embeddings. A comparison of alternative missing-modality imputation strategies is provided in Appendix~\ref{app: all_results}.

\textbf{Baselines.} We benchmark eight dynamic graph methods:
DyGFormer~\cite{yu2023betterdynamicgraphlearning},
TGAT~\cite{xu2020inductiverepresentationlearningtemporal},
GraphMixer~\cite{cong2023reallyneedcomplicatedmodel},
CAWN~\cite{wang2022inductiverepresentationlearningtemporal},
TCL~\cite{wang2021tcltransformerbaseddynamicgraph},
TPNet~\cite{lu2024improving},
SALoM~\cite{liu2025salom},
and FreeDyG~\cite{tian2024freedyg}. In addition, to evaluate the fusion mechanism independently of the temporal backbone, we implement three multimodal fusion strategies on the DyGFormer backbone: gated fusion, modality attention, and MoE fusion~\cite{he2025unigraph}.
Additional details regarding the evaluation framework and experimental settings are provided in Appendix \ref{app:details}.

\subsection{Results}

\begin{figure}
    \centering
    \includegraphics[width=\linewidth]{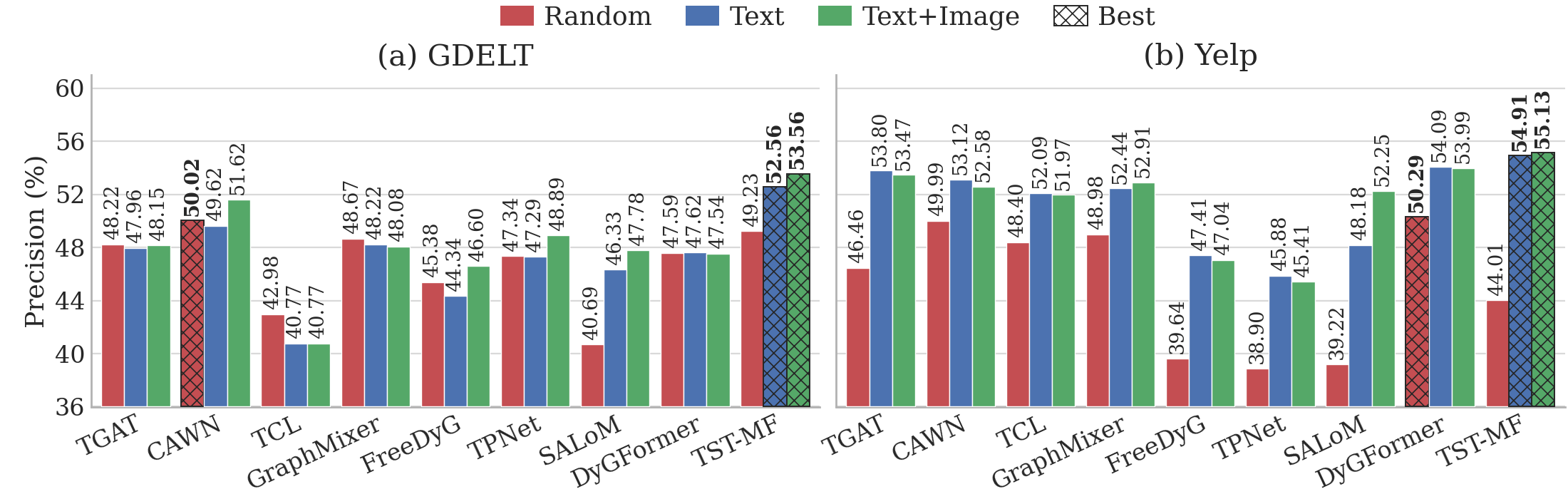}
    \caption{Precision (\%) for edge classification on GDELT and Yelp. The best results are highlighted in shade.}
    \label{fig:edge_classification}
\end{figure}

The performance of the baseline models on the link prediction tasks is shown in Table \ref{tab: overall_result_avg_precision_link_prediction} (and in Appendix \ref{app: all_results} Table \ref{tab: overall_result_auc_link_prediction}). The detailed results on the edge classification task are shown in Figure \ref{fig:edge_classification} (and Appendix \ref{app: all_results} Table \ref{tab: overall_result_avg_precision_edge_classification}, Table \ref{tab: overall_result_avg_recall_edge_classification}, and Table \ref{tab: overall_result_avg_fscore_edge_classification}). Performance on the node retrieval task is shown in Table \ref{tab: overall_result_node_retrieval} (and in Appendix \ref{app: all_results} Table \ref{tab: overall_result_node_retrieval_h1} and Table \ref{tab: overall_result_node_retrieval_h3}), and further assessment on a more challenging setting (historical sampling) are provided in Appendix \ref{app: all_results} Table \ref{tab:supp_recurrence_h10}, Table \ref{tab:supp_recurrence_repeated_h10}, and Table \ref{tab:supp_recurrence_novel_h10}. The bold numbers indicate the best results for each dataset. We observe that:

\textbf{(RQ1) Existing dynamic graph methods work for multimodal situations.} Across the experiments shown in the tables, the results suggest that naive concatenation achieves fairish results. It is often comparable to textual features and occasionally improves retrieval (e.g., DyGFormer on GDELT in Table \ref{tab: overall_result_node_retrieval}). However, the effect is inconsistent, and in some cases multimodal concatenation can even substantially degrade performance (e.g., Figure~\ref{fig:edge_classification} shows that in GDELT, naive concatenation underperforms even a random-feature baseline), motivating more principled fusion.

\begin{figure}[!htbp]
    \centering
    \begin{minipage}[c]{0.5\textwidth} 
        \centering
        \begin{subfigure}[b]{0.48\linewidth}
            \centering
            \includegraphics[width=\linewidth]{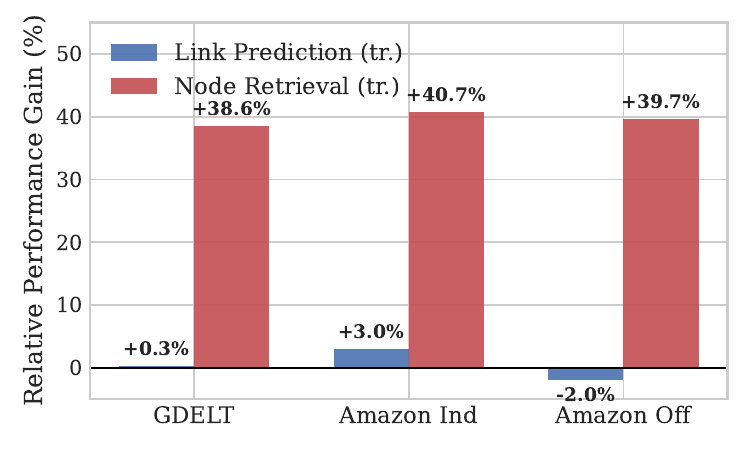}
            \vspace{-15pt}
            \subcaption*{(i)}
        \end{subfigure}
        \hfill
        \begin{subfigure}[b]{0.48\linewidth}
            \centering
            \includegraphics[width=\linewidth]{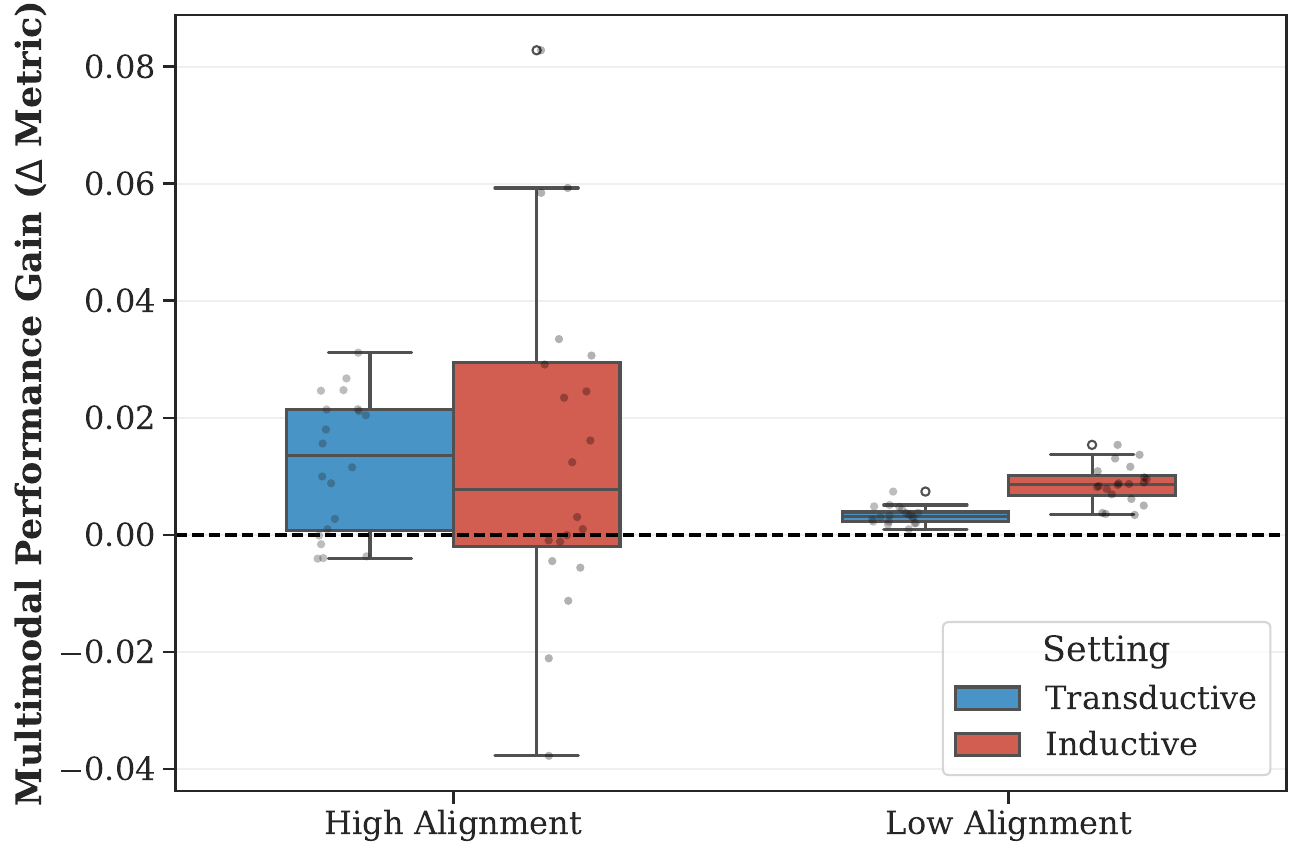}
            \vspace{-15pt}
            \subcaption*{(ii)}
        \end{subfigure}

        \vspace{5pt}

        \begin{subfigure}[b]{1.0\linewidth}
            \centering
            \includegraphics[width=\linewidth]{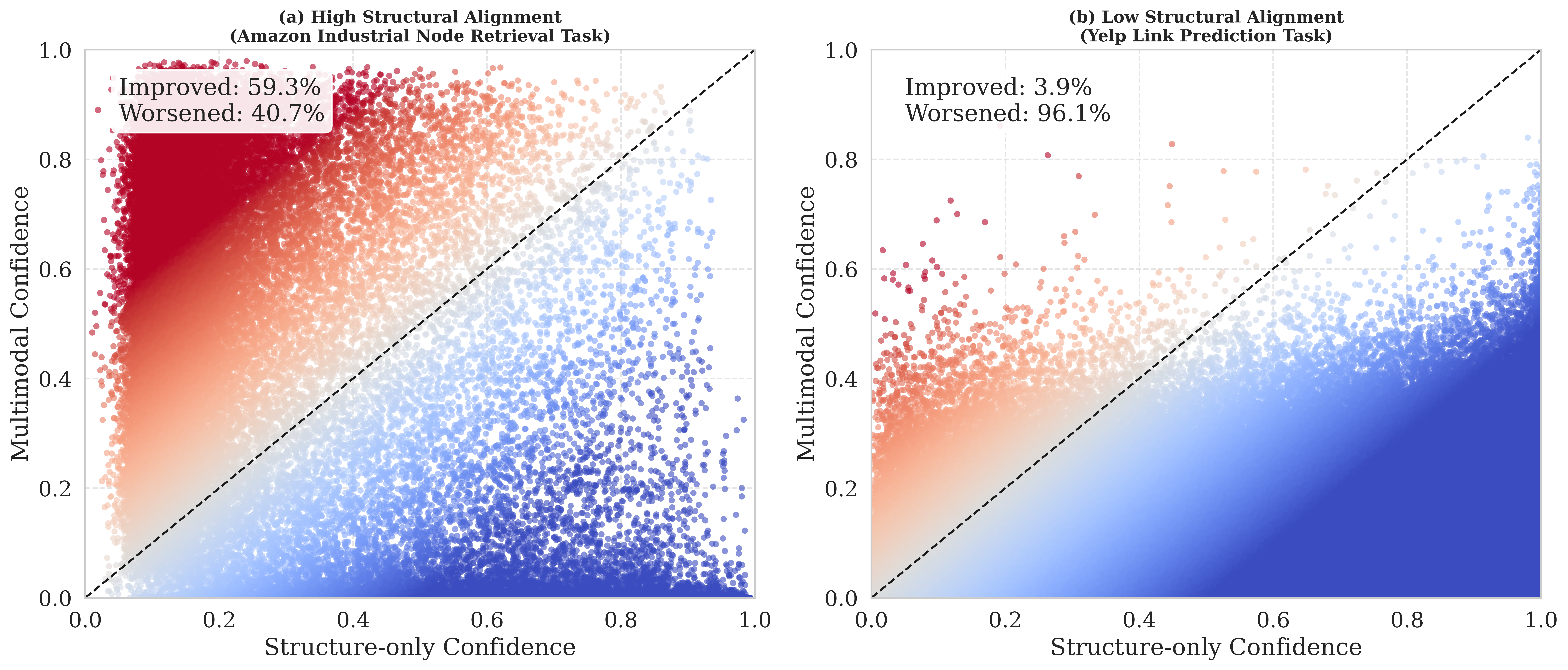}
            \vspace{-15pt}
            \subcaption*{(iii)}
        \end{subfigure}
    \end{minipage}
    \hfill 
    \begin{minipage}[c]{0.48\textwidth} 
        \vspace{0pt} 
        \caption{
            Misaligned modalities can behave as confounders rather than helpful features. (i) Relative performance gains from adding additional modalities over a structure-only baseline across link prediction (transductive) and node retrieval (transductive). Gains are significantly larger for node retrieval than link prediction across most datasets. (ii) Distribution of multimodal performance gains under high vs. low structural alignment, comparing transductive and inductive settings; gains are less reliable under inductive evaluation, where distribution shift can turn visual/text cues into noise. (iii) Instance-level analysis showing multimodal vs. structure-only confidence: points above the diagonal indicate improved confidence, while points below indicate degradation. Even within the same dataset and task, multimodal inputs help some instances but hurt many others. This exposes the risk of indiscriminate modality fusion, especially when task demands and provided modalities are misaligned.
        }
        \label{fig:res_sup_figures}
    \end{minipage}
\end{figure}

\textbf{(RQ2) Additional modalities and additional information do not always guarantee better performance.} Contrary to the intuition that additional information inherently improves representation, the results across the experiments in Table \ref{tab: overall_result_avg_precision_link_prediction}, Table \ref{tab: overall_result_auc_link_prediction}, Figure \ref{fig:edge_classification}, Table \ref{tab: overall_result_node_retrieval} and Figure \ref{fig:res_sup_figures} indicate that the effectiveness of multimodal signals is highly \textbf{dataset- and task-dependent}.

\textbf{First, structural dominance is observed in certain datasets.} In scenarios like Amazon Office and GDELT, strong backbones utilizing \emph{Random embeddings} already achieve high AUC/AP, which suggests that temporal and structural signals dominate the prediction in these contexts and the performance saturation limits the contribution of additional modalities. Consequently, moving from \emph{Random embedding} to \emph{Text} or \emph{Text+Image} yields mixed outcomes. Some methods show modest gains (e.g., GraphMixer on GMCT for link prediction), whereas others show negligible improvements or even degradation (e.g., TCL on GDELT and DyGFormer on Amazon - Office). 

\textbf{Second, there is a distinct task sensitivity regarding multimodal semantics.} Importantly, \emph{we observe a clear task dependence even within the same dataset.} For instance, on both Amazon Office and GDELT, link prediction performance is largely saturated by structural patterns and remains insensitive to multimodal inputs. In contrast, node retrieval, which requires distinguishing specific entity properties, shows a clearer performance gap between \emph{Random} and \emph{Text/Text+Image} features.

Overall, these observations indicate that naively injecting visual signals is not consistently helpful and can introduce noise or misalignment, especially when the visual modality is weakly related to the prediction objective or when the backbone lacks an effective mechanism to integrate multimodal cues with temporal neighborhood evidence.

\textbf{(RQ3) Modality benefits vary from different prediction targets.} While multimodal integration is generally beneficial, we observe that the gains are highly \emph{target-dependent} across \textbf{MM-DyGraph}. We identify the \textbf{structural alignment} between the \emph{visual signal source} and the \emph{prediction target granularity} as the governing factor.

Specifically, tasks benefit most when the visual modality resides directly on the target structure. For instance, Amazon (containing node images) exhibits strong sensitivity in node retrieval tasks (high alignment). Conversely, improvements become inconsistent or negligible when alignment is weak, such as using edge-associated images (from GDELT/GMCT) for node-centric objectives, or node-centric images (from Yelp) for link prediction.

Furthermore, misaligned modalities act as confounders rather than features. Figure~\ref{fig:res_sup_figures} reveals that within the same task and dataset, adding modalities improves some instances while degrading others. This risk is amplified under \emph{inductive settings}, where distribution shifts turn misaligned visual cues into noisy artifacts, degrading performance compared to structure-only baselines. 

\begin{table*}[h]
\centering
\small
\caption{Average Precision (\%) for link prediction. “\textit{tr.}” and “\textit{in.}” refer to the transductive and inductive setting, respectively. The best results are highlighted in bold, and the second-best results are underlined.}
\label{tab: overall_result_avg_precision_link_prediction}
\resizebox{\linewidth}{!}{
\begin{tabular}{c|c|c|ccccccccc}
\toprule
 & \textbf{Datasets} & \textbf{Feature} & \textbf{TGAT} & \textbf{CAWN} & \textbf{TCL} & \textbf{GraphMixer} & \textbf{FreeDyG} & \textbf{TPNet} & \textbf{SALoM} & \textbf{DyGFormer} & \textbf{TST-MF} \\
\midrule
\multirow{15}{*}{\textit{tr.}}

& \multirow{3}{*}{GDELT}
& Random
& 89.15 $\pm$ 1.99 & \underline{92.68 $\pm$ 0.36} & 91.73 $\pm$ 0.02 & 90.83 $\pm$ 0.30 & 90.63 $\pm$ 0.33 & 91.89 $\pm$ 0.58 & 91.21 $\pm$ 1.66 & 92.21 $\pm$ 0.12 & \textbf{92.83 $\pm$ 0.18} \\
& & Text
& 89.54 $\pm$ 1.37 & \underline{92.94 $\pm$ 0.11} & 91.67 $\pm$ 0.16 & 91.65 $\pm$ 0.45 & 91.95 $\pm$ 0.06 & 92.23 $\pm$ 0.19 & 91.64 $\pm$ 0.55 & 92.48 $\pm$ 0.21 & \textbf{94.03 $\pm$ 0.13} \\
& & Text+Image
& 90.76 $\pm$ 0.34 & \underline{93.08 $\pm$ 0.25} & 91.47 $\pm$ 0.19 & 91.38 $\pm$ 0.20 & 91.73 $\pm$ 0.10 & 92.57 $\pm$ 0.36 & 91.89 $\pm$ 0.12 & 92.46 $\pm$ 0.26 & \textbf{93.98 $\pm$ 0.05} \\
\cmidrule{2-12}

& \multirow{3}{*}{GMCT}
& Random
& 74.28 $\pm$ 3.36 & 78.19 $\pm$ 0.05 & 78.07 $\pm$ 0.17 & 78.26 $\pm$ 0.05 & 78.28 $\pm$ 0.11 & 79.53 $\pm$ 0.43 & 62.47 $\pm$ 1.35 & \textbf{94.47 $\pm$ 0.78} & \underline{80.79 $\pm$ 3.27} \\
& & Text
& 82.04 $\pm$ 3.67 & 79.49 $\pm$ 0.09 & 78.55 $\pm$ 0.34 & 79.26 $\pm$ 0.30 & 78.88 $\pm$ 0.29 & 79.81 $\pm$ 0.29 & 60.96 $\pm$ 1.25 & \textbf{95.73 $\pm$ 0.30} & \underline{89.77 $\pm$ 0.65} \\
& & Text+Image
& 81.62 $\pm$ 2.33 & 79.65 $\pm$ 0.05 & 78.50 $\pm$ 0.22 & 79.62 $\pm$ 0.48 & 78.93 $\pm$ 0.19 & 79.83 $\pm$ 0.26 & 60.04 $\pm$ 3.08 & \textbf{95.54 $\pm$ 0.10} & \underline{89.08 $\pm$ 0.72} \\
\cmidrule{2-12}

& \multirow{3}{*}{Amazon - Industrial}
& Random
& 91.22 $\pm$ 0.77 & 91.08 $\pm$ 0.19 & 91.31 $\pm$ 0.53 & \underline{92.27 $\pm$ 0.23} & 90.55 $\pm$ 0.81 & 91.73 $\pm$ 0.79 & 82.22 $\pm$ 2.85 & 89.88 $\pm$ 0.73 & \textbf{92.44 $\pm$ 0.19} \\
& & Text
& 93.74 $\pm$ 0.13 & 91.94 $\pm$ 0.03 & 93.11 $\pm$ 0.51 & \textbf{94.10 $\pm$ 0.11} & 92.25 $\pm$ 0.57 & 92.52 $\pm$ 0.18 & 83.33 $\pm$ 1.97 & 92.45 $\pm$ 0.31 & \underline{94.04 $\pm$ 0.22} \\
& & Text+Image
& \underline{93.97 $\pm$ 0.18} & 92.21 $\pm$ 0.18 & 93.11 $\pm$ 0.36 & 93.96 $\pm$ 0.37 & 91.95 $\pm$ 0.40 & 92.91 $\pm$ 0.51 & 83.64 $\pm$ 1.67 & 92.56 $\pm$ 0.24 & \textbf{94.10 $\pm$ 0.12} \\
\cmidrule{2-12}

& \multirow{3}{*}{Amazon - Office}
& Random
& 90.51 $\pm$ 1.72 & 89.28 $\pm$ 0.56 & \textbf{91.87 $\pm$ 0.30} & 83.53 $\pm$ 0.08 & 89.04 $\pm$ 1.13 & 89.90 $\pm$ 1.42 & 87.32 $\pm$ 1.70 & 90.82 $\pm$ 0.76 & \underline{91.27 $\pm$ 0.20} \\
& & Text
& \textbf{92.11 $\pm$ 1.25} & 91.06 $\pm$ 0.41 & 91.36 $\pm$ 0.38 & 90.25 $\pm$ 2.46 & 91.78 $\pm$ 0.27 & 90.72 $\pm$ 1.69 & 87.70 $\pm$ 1.21 & 90.37 $\pm$ 1.98 & \underline{92.04 $\pm$ 0.12} \\
& & Text+Image
& 91.23 $\pm$ 0.23 & 91.64 $\pm$ 0.29 & 91.73 $\pm$ 0.19 & 91.06 $\pm$ 2.56 & \underline{91.74 $\pm$ 0.72} & 89.90 $\pm$ 0.34 & 86.21 $\pm$ 1.74 & 89.00 $\pm$ 0.23 & \textbf{92.67 $\pm$ 0.10} \\
\cmidrule{2-12}

& \multirow{3}{*}{Yelp}
& Random
& 80.91 $\pm$ 0.16 & 80.54 $\pm$ 0.13 & 80.31 $\pm$ 0.69 & 80.43 $\pm$ 0.52 & 81.90 $\pm$ 0.07 & \textbf{91.08 $\pm$ 0.06} & 79.30 $\pm$ 0.97 & 81.83 $\pm$ 0.07 & \underline{82.11 $\pm$ 0.05} \\
& & Text
& \underline{92.42 $\pm$ 0.16} & 92.27 $\pm$ 0.19 & 89.71 $\pm$ 1.96 & 87.32 $\pm$ 0.07 & 90.82 $\pm$ 0.08 & 91.49 $\pm$ 0.10 & 80.98 $\pm$ 0.54 & 88.75 $\pm$ 5.83 & \textbf{92.61 $\pm$ 0.02} \\
& & Text+Image
& \underline{92.22 $\pm$ 0.42} & 91.94 $\pm$ 0.13 & 87.44 $\pm$ 5.74 & 87.26 $\pm$ 0.19 & 90.61 $\pm$ 0.10 & 91.76 $\pm$ 0.20 & 79.02 $\pm$ 0.75 & 88.55 $\pm$ 5.51 & \textbf{92.65 $\pm$ 0.04} \\
\midrule

\multirow{15}{*}{\textit{in.}}

& \multirow{3}{*}{GDELT}
& Random
& 76.71 $\pm$ 0.43 & 79.65 $\pm$ 0.35 & 78.37 $\pm$ 0.10 & 77.37 $\pm$ 1.03 & 77.35 $\pm$ 0.44 & 77.41 $\pm$ 1.67 & 79.29 $\pm$ 0.27 & \textbf{81.70 $\pm$ 0.30} & \underline{81.14 $\pm$ 0.36} \\
& & Text
& 76.91 $\pm$ 1.53 & 80.87 $\pm$ 0.47 & 78.46 $\pm$ 0.23 & 78.84 $\pm$ 0.61 & 80.71 $\pm$ 0.21 & 78.30 $\pm$ 0.20 & 79.40 $\pm$ 0.43 & \underline{83.07 $\pm$ 0.28} & \textbf{85.35 $\pm$ 0.17} \\
& & Text+Image
& 78.38 $\pm$ 0.67 & 81.02 $\pm$ 0.41 & 78.33 $\pm$ 0.29 & 78.58 $\pm$ 0.34 & 80.33 $\pm$ 0.21 & 79.68 $\pm$ 1.05 & 80.03 $\pm$ 0.58 & \underline{83.05 $\pm$ 0.28} & \textbf{85.44 $\pm$ 0.14} \\
\cmidrule{2-12}

& \multirow{3}{*}{GMCT}
& Random
& 60.42 $\pm$ 2.40 & 61.08 $\pm$ 0.19 & 61.11 $\pm$ 0.23 & 61.26 $\pm$ 0.14 & 61.27 $\pm$ 0.39 & \underline{63.32 $\pm$ 0.37} & 52.88 $\pm$ 0.69 & \textbf{79.51 $\pm$ 1.22} & 62.26 $\pm$ 0.99 \\
& & Text
& \underline{67.10 $\pm$ 6.29} & 62.12 $\pm$ 0.14 & 61.27 $\pm$ 0.16 & 62.23 $\pm$ 0.68 & 61.72 $\pm$ 0.44 & 62.93 $\pm$ 0.19 & 52.28 $\pm$ 0.15 & \textbf{83.46 $\pm$ 2.84} & 64.90 $\pm$ 1.33 \\
& & Text+Image
& \underline{66.02 $\pm$ 4.08} & 62.30 $\pm$ 0.18 & 60.90 $\pm$ 0.23 & 62.83 $\pm$ 0.89 & 61.71 $\pm$ 0.37 & 63.28 $\pm$ 0.45 & 51.99 $\pm$ 1.07 & \textbf{83.05 $\pm$ 3.36} & 65.49 $\pm$ 1.00 \\
\cmidrule{2-12}

& \multirow{3}{*}{Amazon - Industrial}
& Random
& 74.73 $\pm$ 2.86 & 74.30 $\pm$ 0.67 & 74.92 $\pm$ 2.04 & \underline{78.07 $\pm$ 0.66} & 72.30 $\pm$ 2.66 & 75.81 $\pm$ 2.67 & 65.67 $\pm$ 2.24 & 74.44 $\pm$ 0.24 & \textbf{78.72 $\pm$ 0.72} \\
& & Text
& 81.58 $\pm$ 0.42 & 75.89 $\pm$ 0.11 & 79.60 $\pm$ 1.88 & \textbf{83.15 $\pm$ 0.39} & 77.02 $\pm$ 1.83 & 77.93 $\pm$ 0.64 & 65.86 $\pm$ 2.09 & 76.47 $\pm$ 0.69 & \underline{81.67 $\pm$ 0.77} \\
& & Text+Image
& \underline{82.29 $\pm$ 0.70} & 76.74 $\pm$ 0.86 & 79.70 $\pm$ 1.09 & \textbf{82.75 $\pm$ 1.15} & 75.90 $\pm$ 1.29 & 79.32 $\pm$ 1.63 & 66.03 $\pm$ 1.63 & 76.33 $\pm$ 0.58 & 81.95 $\pm$ 0.54 \\
\cmidrule{2-12}

& \multirow{3}{*}{Amazon - Office}
& Random
& 74.14 $\pm$ 4.50 & 70.12 $\pm$ 1.59 & \textbf{77.21 $\pm$ 0.74} & 57.70 $\pm$ 0.10 & 69.19 $\pm$ 3.44 & 71.87 $\pm$ 4.13 & 71.87 $\pm$ 1.84 & 75.77 $\pm$ 1.74 & \underline{75.91 $\pm$ 0.32} \\
& & Text
& \textbf{78.07 $\pm$ 4.00} & 74.62 $\pm$ 1.36 & 75.78 $\pm$ 1.31 & 72.29 $\pm$ 7.16 & 76.78 $\pm$ 0.67 & 73.65 $\pm$ 5.06 & 71.21 $\pm$ 1.47 & 73.50 $\pm$ 6.60 & \underline{77.47 $\pm$ 0.29} \\
& & Text+Image
& 74.85 $\pm$ 0.80 & \underline{76.70 $\pm$ 0.84} & 76.61 $\pm$ 0.54 & 75.02 $\pm$ 7.04 & 76.60 $\pm$ 2.16 & 71.25 $\pm$ 0.97 & 71.16 $\pm$ 2.10 & 68.64 $\pm$ 1.02 & \textbf{78.16 $\pm$ 0.31} \\
\cmidrule{2-12}

& \multirow{3}{*}{Yelp}
& Random
& 77.46 $\pm$ 0.22 & 76.57 $\pm$ 0.37 & 76.93 $\pm$ 0.30 & 76.83 $\pm$ 0.47 & 78.21 $\pm$ 0.09 & \textbf{86.97 $\pm$ 0.10} & 72.68 $\pm$ 0.67 & 77.99 $\pm$ 0.09 & \underline{78.33 $\pm$ 0.06} \\
& & Text
& \underline{89.09 $\pm$ 0.20} & 88.81 $\pm$ 0.25 & 86.05 $\pm$ 2.08 & 83.43 $\pm$ 0.13 & 87.23 $\pm$ 0.09 & 87.48 $\pm$ 0.08 & 74.02 $\pm$ 0.74 & 85.29 $\pm$ 6.09 & \textbf{89.32 $\pm$ 0.04} \\
& & Text+Image
& \underline{88.93 $\pm$ 0.53} & 88.38 $\pm$ 0.20 & 83.91 $\pm$ 5.83 & 83.45 $\pm$ 0.16 & 86.98 $\pm$ 0.13 & 87.83 $\pm$ 0.25 & 71.63 $\pm$ 0.46 & 85.05 $\pm$ 5.64 & \textbf{89.39 $\pm$ 0.05} \\
\bottomrule
\end{tabular}}
\end{table*}

\begin{table*}[h]
\centering
\small
\caption{Hits@10 (\%) for node retrieval. “\textit{tr.}” and “\textit{in.}” refer to the transductive and inductive setting, respectively. The best results are highlighted in bold, and the second-best results are underlined.}
\label{tab: overall_result_node_retrieval}
\resizebox{\linewidth}{!}{
\begin{tabular}{c|c|c|ccccccccc}
\toprule
 & \textbf{Datasets} & \textbf{Feature} & \textbf{TGAT} & \textbf{CAWN} & \textbf{TCL} & \textbf{GraphMixer} & \textbf{FreeDyG} & \textbf{TPNet} & \textbf{SALoM} & \textbf{DyGFormer} & \textbf{TST-MF} \\
\midrule
\multirow{15}{*}{\textit{tr.}}

& \multirow{3}{*}{GDELT}
& Random
& 67.77 $\pm$ 3.43 & \underline{77.54 $\pm$ 0.69} & 75.42 $\pm$ 0.43 & 70.03 $\pm$ 0.99 & 69.38 $\pm$ 1.71 & 74.73 $\pm$ 1.55 & 73.62 $\pm$ 5.37 & 52.96 $\pm$ 4.82 & \textbf{77.72 $\pm$ 0.36} \\
& & Text
& 67.28 $\pm$ 3.85 & \underline{78.81 $\pm$ 0.43} & 74.99 $\pm$ 0.42 & 72.45 $\pm$ 1.71 & 76.47 $\pm$ 0.66 & 75.57 $\pm$ 0.61 & 73.99 $\pm$ 2.02 & 69.78 $\pm$ 8.64 & \textbf{81.21 $\pm$ 0.78} \\
& & Text+Image
& 70.79 $\pm$ 1.04 & \underline{79.08 $\pm$ 0.35} & 74.66 $\pm$ 0.43 & 72.41 $\pm$ 1.16 & 74.09 $\pm$ 1.77 & 76.21 $\pm$ 0.61 & 75.02 $\pm$ 0.21 & 71.34 $\pm$ 7.82 & \textbf{81.62 $\pm$ 0.33} \\
\cmidrule{2-12}

& \multirow{3}{*}{GMCT}
& Random
& 42.35 $\pm$ 5.98 & 50.29 $\pm$ 0.20 & 50.13 $\pm$ 0.30 & 50.22 $\pm$ 0.12 & 50.90 $\pm$ 0.19 & 50.79 $\pm$ 0.51 & 20.74 $\pm$ 0.06 & \textbf{100.00 $\pm$ 0.00} & \underline{78.77 $\pm$ 24.87} \\
& & Text
& \underline{54.01 $\pm$ 5.80} & 50.57 $\pm$ 0.11 & 50.34 $\pm$ 0.08 & 50.14 $\pm$ 0.41 & 51.21 $\pm$ 0.30 & 51.19 $\pm$ 0.24 & 20.89 $\pm$ 0.18 & \textbf{100.00 $\pm$ 0.00} & \textbf{100.00 $\pm$ 0.00} \\
& & Text+Image
& \underline{53.02 $\pm$ 2.88} & 50.73 $\pm$ 0.02 & 50.27 $\pm$ 0.24 & 50.72 $\pm$ 0.69 & 51.18 $\pm$ 0.11 & 51.24 $\pm$ 0.23 & 20.69 $\pm$ 0.22 & \textbf{100.00 $\pm$ 0.00} & \textbf{100.00 $\pm$ 0.00} \\
\cmidrule{2-12}

& \multirow{3}{*}{Amazon - Industrial}
& Random
& 73.68 $\pm$ 0.80 & 73.23 $\pm$ 0.79 & 73.43 $\pm$ 0.91 & 75.47 $\pm$ 0.29 & 73.81 $\pm$ 0.58 & 74.79 $\pm$ 0.91 & 56.84 $\pm$ 3.80 & \textbf{99.51 $\pm$ 0.79} & \underline{85.82 $\pm$ 9.03} \\
& & Text
& 79.40 $\pm$ 0.32 & 76.96 $\pm$ 0.38 & 79.02 $\pm$ 0.40 & \underline{80.47 $\pm$ 0.09} & 76.78 $\pm$ 0.84 & 76.76 $\pm$ 0.41 & 58.62 $\pm$ 3.24 & \textbf{100.00 $\pm$ 0.01} & \textbf{100.00 $\pm$ 0.00} \\
& & Text+Image
& \underline{79.92 $\pm$ 0.13} & 77.07 $\pm$ 0.57 & 78.61 $\pm$ 1.00 & 79.72 $\pm$ 1.43 & 76.29 $\pm$ 0.44 & 77.31 $\pm$ 0.74 & 61.84 $\pm$ 3.00 & \textbf{100.00 $\pm$ 0.00} & \textbf{100.00 $\pm$ 0.00} \\
\cmidrule{2-12}

& \multirow{3}{*}{Amazon - Office}
& Random
& 69.70 $\pm$ 3.29 & 67.33 $\pm$ 2.40 & 72.87 $\pm$ 0.95 & 66.83 $\pm$ 0.08 & 69.23 $\pm$ 1.05 & 70.64 $\pm$ 1.43 & 65.74 $\pm$ 2.27 & \textbf{99.78 $\pm$ 0.15} & \underline{74.64 $\pm$ 3.84} \\
& & Text
& 74.33 $\pm$ 2.43 & 72.25 $\pm$ 1.16 & 71.68 $\pm$ 1.17 & 72.51 $\pm$ 3.38 & 73.68 $\pm$ 0.54 & 72.02 $\pm$ 2.07 & 67.13 $\pm$ 1.86 & \underline{99.49 $\pm$ 0.63} & \textbf{99.71 $\pm$ 0.34} \\
& & Text+Image
& 73.38 $\pm$ 0.47 & 72.72 $\pm$ 1.24 & 73.01 $\pm$ 0.94 & 73.42 $\pm$ 3.39 & \underline{73.80 $\pm$ 1.22} & 70.30 $\pm$ 0.48 & 64.35 $\pm$ 2.94 & \textbf{99.99 $\pm$ 0.01} & \textbf{99.99 $\pm$ 0.01} \\
\cmidrule{2-12}

& \multirow{3}{*}{Yelp}
& Random
& 48.18 $\pm$ 0.35 & 46.80 $\pm$ 0.19 & 46.91 $\pm$ 1.38 & 46.45 $\pm$ 1.52 & 49.86 $\pm$ 0.29 & \textbf{71.57 $\pm$ 0.16} & \underline{52.60 $\pm$ 1.29} & 25.99 $\pm$ 14.39 & 50.04 $\pm$ 0.38 \\
& & Text
& \underline{74.59 $\pm$ 0.38} & 73.91 $\pm$ 0.50 & 67.97 $\pm$ 4.75 & 61.89 $\pm$ 0.28 & 70.87 $\pm$ 0.26 & 72.22 $\pm$ 0.06 & 55.17 $\pm$ 1.61 & 67.41 $\pm$ 16.19 & \textbf{75.19 $\pm$ 1.51} \\
& & Text+Image
& \underline{74.19 $\pm$ 1.22} & 72.94 $\pm$ 0.44 & 63.01 $\pm$ 13.26 & 61.72 $\pm$ 0.29 & 70.14 $\pm$ 0.36 & 72.80 $\pm$ 0.39 & 50.75 $\pm$ 1.58 & 67.54 $\pm$ 14.55 & \textbf{79.11 $\pm$ 0.49} \\
\midrule

\multirow{15}{*}{\textit{in.}}

& \multirow{3}{*}{GDELT}
& Random
& 44.74 $\pm$ 1.32 & 47.36 $\pm$ 0.85 & 48.49 $\pm$ 0.18 & 44.83 $\pm$ 4.32 & 45.97 $\pm$ 0.73 & 43.84 $\pm$ 3.19 & \underline{49.30 $\pm$ 0.62} & 33.75 $\pm$ 7.39 & \textbf{50.38 $\pm$ 0.89} \\
& & Text
& 45.41 $\pm$ 3.31 & 49.50 $\pm$ 0.94 & 48.48 $\pm$ 0.35 & 48.54 $\pm$ 1.07 & \underline{51.95 $\pm$ 0.92} & 46.61 $\pm$ 0.61 & 48.87 $\pm$ 1.42 & 51.41 $\pm$ 5.13 & \textbf{58.87 $\pm$ 0.63} \\
& & Text+Image
& 46.86 $\pm$ 1.00 & 49.69 $\pm$ 0.94 & 48.55 $\pm$ 0.27 & 47.72 $\pm$ 1.15 & 51.29 $\pm$ 0.38 & 47.48 $\pm$ 1.80 & 50.18 $\pm$ 0.61 & \underline{52.50 $\pm$ 3.86} & \textbf{59.23 $\pm$ 0.58} \\
\cmidrule{2-12}

& \multirow{3}{*}{GMCT}
& Random
& 20.64 $\pm$ 2.60 & 22.50 $\pm$ 0.17 & 22.64 $\pm$ 0.18 & 23.09 $\pm$ 0.24 & 41.63 $\pm$ 0.89 & 24.07 $\pm$ 0.23 & 12.59 $\pm$ 0.09 & \textbf{100.00 $\pm$ 0.00} & \underline{67.55 $\pm$ 27.64} \\
& & Text
& 29.17 $\pm$ 7.33 & 22.81 $\pm$ 0.26 & 22.63 $\pm$ 0.39 & 21.66 $\pm$ 1.49 & \underline{41.06 $\pm$ 0.34} & 23.69 $\pm$ 0.21 & 12.55 $\pm$ 0.03 & \textbf{100.00 $\pm$ 0.00} & \textbf{100.00 $\pm$ 0.00} \\
& & Text+Image
& 27.40 $\pm$ 4.31 & 23.23 $\pm$ 0.12 & 22.75 $\pm$ 0.33 & 23.15 $\pm$ 0.75 & \underline{40.95 $\pm$ 0.18} & 24.42 $\pm$ 0.12 & 12.64 $\pm$ 0.14 & \textbf{100.00 $\pm$ 0.01} & \textbf{100.00 $\pm$ 0.00} \\
\cmidrule{2-12}

& \multirow{3}{*}{Amazon - Industrial}
& Random
& 41.04 $\pm$ 5.86 & 41.28 $\pm$ 0.93 & 42.23 $\pm$ 6.18 & 47.93 $\pm$ 0.77 & 45.23 $\pm$ 4.40 & 40.94 $\pm$ 5.67 & 28.05 $\pm$ 4.80 & \textbf{78.07 $\pm$ 30.19} & \underline{58.66 $\pm$ 6.40} \\
& & Text
& 51.76 $\pm$ 1.26 & 39.12 $\pm$ 0.04 & 48.24 $\pm$ 3.21 & 54.64 $\pm$ 0.58 & 52.68 $\pm$ 4.78 & 45.41 $\pm$ 1.87 & 27.60 $\pm$ 1.93 & \textbf{99.23 $\pm$ 0.19} & \underline{97.79 $\pm$ 0.35} \\
& & Text+Image
& 53.98 $\pm$ 1.23 & 41.28 $\pm$ 1.89 & 48.54 $\pm$ 2.34 & 54.10 $\pm$ 1.97 & 50.49 $\pm$ 1.85 & 48.08 $\pm$ 4.19 & 29.58 $\pm$ 4.71 & \underline{99.51 $\pm$ 0.19} & \textbf{99.94 $\pm$ 0.99} \\
\cmidrule{2-12}

& \multirow{3}{*}{Amazon - Office}
& Random
& 40.20 $\pm$ 11.05 & 31.02 $\pm$ 2.38 & \underline{48.22 $\pm$ 1.86} & 15.08 $\pm$ 1.54 & 36.86 $\pm$ 5.76 & 34.18 $\pm$ 8.18 & 37.27 $\pm$ 3.74 & \textbf{83.63 $\pm$ 11.06} & 46.55 $\pm$ 1.70 \\
& & Text
& 45.50 $\pm$ 9.16 & 38.60 $\pm$ 3.86 & 43.91 $\pm$ 3.72 & 34.02 $\pm$ 15.24 & 52.76 $\pm$ 1.60 & 37.67 $\pm$ 10.66 & 37.31 $\pm$ 1.80 & \textbf{96.83 $\pm$ 0.32} & \underline{95.65 $\pm$ 8.96} \\
& & Text+Image
& 38.04 $\pm$ 1.99 & 43.72 $\pm$ 2.65 & 46.78 $\pm$ 1.55 & 40.61 $\pm$ 11.80 & 52.58 $\pm$ 5.56 & 31.96 $\pm$ 3.12 & 36.86 $\pm$ 3.21 & \underline{98.10 $\pm$ 0.61} & \textbf{98.99 $\pm$ 0.31} \\
\cmidrule{2-12}

& \multirow{3}{*}{Yelp}
& Random
& 43.14 $\pm$ 0.26 & 41.60 $\pm$ 0.36 & 42.30 $\pm$ 0.61 & 42.00 $\pm$ 1.02 & 44.58 $\pm$ 0.13 & \textbf{61.85 $\pm$ 0.10} & 41.13 $\pm$ 0.91 & 26.12 $\pm$ 9.72 & \underline{44.77 $\pm$ 0.31} \\
& & Text
& \underline{65.50 $\pm$ 0.46} & 64.53 $\pm$ 0.61 & 58.46 $\pm$ 4.39 & 53.68 $\pm$ 0.25 & 61.75 $\pm$ 0.23 & 62.55 $\pm$ 0.22 & 42.92 $\pm$ 1.67 & 59.47 $\pm$ 14.07 & \textbf{66.84 $\pm$ 0.38} \\
& & Text+Image
& \underline{65.24 $\pm$ 1.17} & 63.49 $\pm$ 0.58 & 54.84 $\pm$ 10.96 & 53.71 $\pm$ 0.20 & 61.22 $\pm$ 0.26 & 63.28 $\pm$ 0.30 & 38.81 $\pm$ 0.86 & 59.48 $\pm$ 12.06 & \textbf{67.84 $\pm$ 0.48} \\
\bottomrule
\end{tabular}}
\end{table*}

\begin{table*}[h]
\centering
\small
\caption{Comparison of multimodal fusion strategies under the Text+Image setting (\%). All methods share the DyGFormer backbone, and DyGFormer denotes naive feature concatenation. “\textit{tr.}” and “\textit{in.}” refer to the transductive and inductive setting, respectively. The best results are highlighted in bold, and the second-best results are underlined.}
\label{tab:fusion_comparison}
\resizebox{\linewidth}{!}{
\begin{tabular}{c|c|c|c|c|ccc|c}
\toprule
\textbf{Task} & \textbf{Metric} & & \textbf{Datasets} & \textbf{DyGFormer} & \textbf{Gated fusion} & \textbf{Modality attention} & \textbf{MoE fusion} & \textbf{TST-MF} \\
\midrule
\multirow{10}{*}{LP} & \multirow{10}{*}{AP} & \multirow{5}{*}{\textit{tr.}} & GDELT & 92.46 $\pm$ 0.26 & 92.74 $\pm$ 0.28 & \underline{93.08 $\pm$ 0.10} & 92.92 $\pm$ 0.26 & \textbf{93.98 $\pm$ 0.05} \\
 &  &  & GMCT & 95.54 $\pm$ 0.10 & \underline{96.41 $\pm$ 0.17} & 96.30 $\pm$ 0.20 & \textbf{96.59 $\pm$ 0.16} & 89.08 $\pm$ 0.72 \\
 &  &  & Amazon - Industrial & 92.56 $\pm$ 0.24 & \underline{92.63 $\pm$ 0.26} & 92.61 $\pm$ 0.61 & 90.89 $\pm$ 2.25 & \textbf{94.10 $\pm$ 0.12} \\
 &  &  & Amazon - Office & 89.00 $\pm$ 0.23 & 90.26 $\pm$ 1.41 & \underline{90.73 $\pm$ 1.02} & 89.59 $\pm$ 0.94 & \textbf{92.67 $\pm$ 0.10} \\
 &  &  & Yelp & 88.55 $\pm$ 5.51 & 91.25 $\pm$ 0.38 & \underline{92.29 $\pm$ 0.17} & 88.56 $\pm$ 5.75 & \textbf{92.65 $\pm$ 0.04} \\
\cmidrule{3-9}
 &  & \multirow{5}{*}{\textit{in.}} & GDELT & 83.05 $\pm$ 0.28 & 83.29 $\pm$ 0.34 & 84.07 $\pm$ 0.31 & \underline{84.14 $\pm$ 0.35} & \textbf{85.44 $\pm$ 0.14} \\
 &  &  & GMCT & \textbf{83.05 $\pm$ 3.36} & \underline{82.46 $\pm$ 1.83} & 81.92 $\pm$ 2.71 & 82.28 $\pm$ 1.28 & 65.49 $\pm$ 1.00 \\
 &  &  & Amazon - Industrial & 76.33 $\pm$ 0.58 & \underline{77.35 $\pm$ 0.83} & 76.44 $\pm$ 2.17 & 74.98 $\pm$ 1.07 & \textbf{81.95 $\pm$ 0.54} \\
 &  &  & Amazon - Office & 68.64 $\pm$ 1.02 & 72.38 $\pm$ 4.30 & \underline{74.01 $\pm$ 3.95} & 70.59 $\pm$ 3.07 & \textbf{78.16 $\pm$ 0.31} \\
 &  &  & Yelp & 85.05 $\pm$ 5.64 & 87.73 $\pm$ 0.35 & \underline{88.96 $\pm$ 0.21} & 85.00 $\pm$ 5.95 & \textbf{89.39 $\pm$ 0.05} \\
\midrule
\multirow{10}{*}{NR} & \multirow{10}{*}{Hits@10} & \multirow{5}{*}{\textit{tr.}} & GDELT & 71.34 $\pm$ 7.82 & 62.18 $\pm$ 6.57 & \underline{74.46 $\pm$ 1.58} & 71.33 $\pm$ 2.94 & \textbf{81.62 $\pm$ 0.33} \\
 &  &  & GMCT & \textbf{100.00 $\pm$ 0.00} & \textbf{100.00 $\pm$ 0.00} & \textbf{100.00 $\pm$ 0.00} & \underline{99.96 $\pm$ 0.07} & \textbf{100.00 $\pm$ 0.00} \\
 &  &  & Amazon - Industrial & \textbf{100.00 $\pm$ 0.00} & \textbf{100.00 $\pm$ 0.00} & \textbf{100.00 $\pm$ 0.01} & \underline{99.66 $\pm$ 0.58} & \textbf{100.00 $\pm$ 0.00} \\
 &  &  & Amazon - Office & \textbf{99.99 $\pm$ 0.01} & 99.25 $\pm$ 1.27 & \underline{99.98 $\pm$ 0.03} & 98.45 $\pm$ 1.52 & \textbf{99.99 $\pm$ 0.01} \\
 &  &  & Yelp & 67.54 $\pm$ 14.55 & \underline{75.02 $\pm$ 1.72} & 57.09 $\pm$ 35.30 & 66.89 $\pm$ 17.38 & \textbf{79.11 $\pm$ 0.49} \\
\cmidrule{3-9}
 &  & \multirow{5}{*}{\textit{in.}} & GDELT & 52.50 $\pm$ 3.86 & 46.75 $\pm$ 7.15 & \underline{54.62 $\pm$ 0.68} & 54.57 $\pm$ 1.51 & \textbf{59.23 $\pm$ 0.58} \\
 &  &  & GMCT & \textbf{100.00 $\pm$ 0.01} & \textbf{100.00 $\pm$ 0.00} & \textbf{100.00 $\pm$ 0.00} & \underline{99.88 $\pm$ 0.21} & \textbf{100.00 $\pm$ 0.00} \\
 &  &  & Amazon - Industrial & \underline{99.51 $\pm$ 0.19} & 98.25 $\pm$ 1.59 & 98.65 $\pm$ 0.86 & 96.51 $\pm$ 4.64 & \textbf{99.94 $\pm$ 0.99} \\
 &  &  & Amazon - Office & \underline{98.10 $\pm$ 0.61} & 96.58 $\pm$ 1.56 & 96.36 $\pm$ 1.14 & 95.89 $\pm$ 2.37 & \textbf{98.99 $\pm$ 0.31} \\
 &  &  & Yelp & 59.48 $\pm$ 12.06 & \underline{65.29 $\pm$ 1.92} & 55.40 $\pm$ 22.19 & 58.69 $\pm$ 15.22 & \textbf{67.84 $\pm$ 0.48} \\
\midrule
\multirow{5}{*}{EC} & \multirow{5}{*}{Precision} &  & GDELT & 47.54 $\pm$ 0.10 & 48.49 $\pm$ 0.71 & \underline{48.85 $\pm$ 0.89} & 47.74 $\pm$ 0.25 & \textbf{53.56 $\pm$ 1.70} \\
 &  &  & GMCT & \underline{56.77 $\pm$ 1.72} & \textbf{57.69 $\pm$ 6.60} & 56.36 $\pm$ 3.38 & 54.81 $\pm$ 2.50 & 53.45 $\pm$ 1.57 \\
 &  &  & Amazon - Industrial & \textbf{59.38 $\pm$ 1.85} & 52.81 $\pm$ 0.68 & 52.53 $\pm$ 0.57 & \underline{53.50 $\pm$ 0.20} & 50.32 $\pm$ 1.86 \\
 &  &  & Amazon - Office & \textbf{56.02 $\pm$ 0.50} & 53.98 $\pm$ 0.86 & 53.73 $\pm$ 0.12 & 53.91 $\pm$ 0.44 & \underline{54.42 $\pm$ 0.02} \\
 &  &  & Yelp & \underline{53.99 $\pm$ 0.18} & 52.65 $\pm$ 0.30 & 53.91 $\pm$ 0.49 & 52.71 $\pm$ 1.03 & \textbf{55.13 $\pm$ 0.87} \\
\bottomrule
\end{tabular}}
\end{table*}

\begin{figure}[h]
    \centering
    \includegraphics[width=\linewidth]{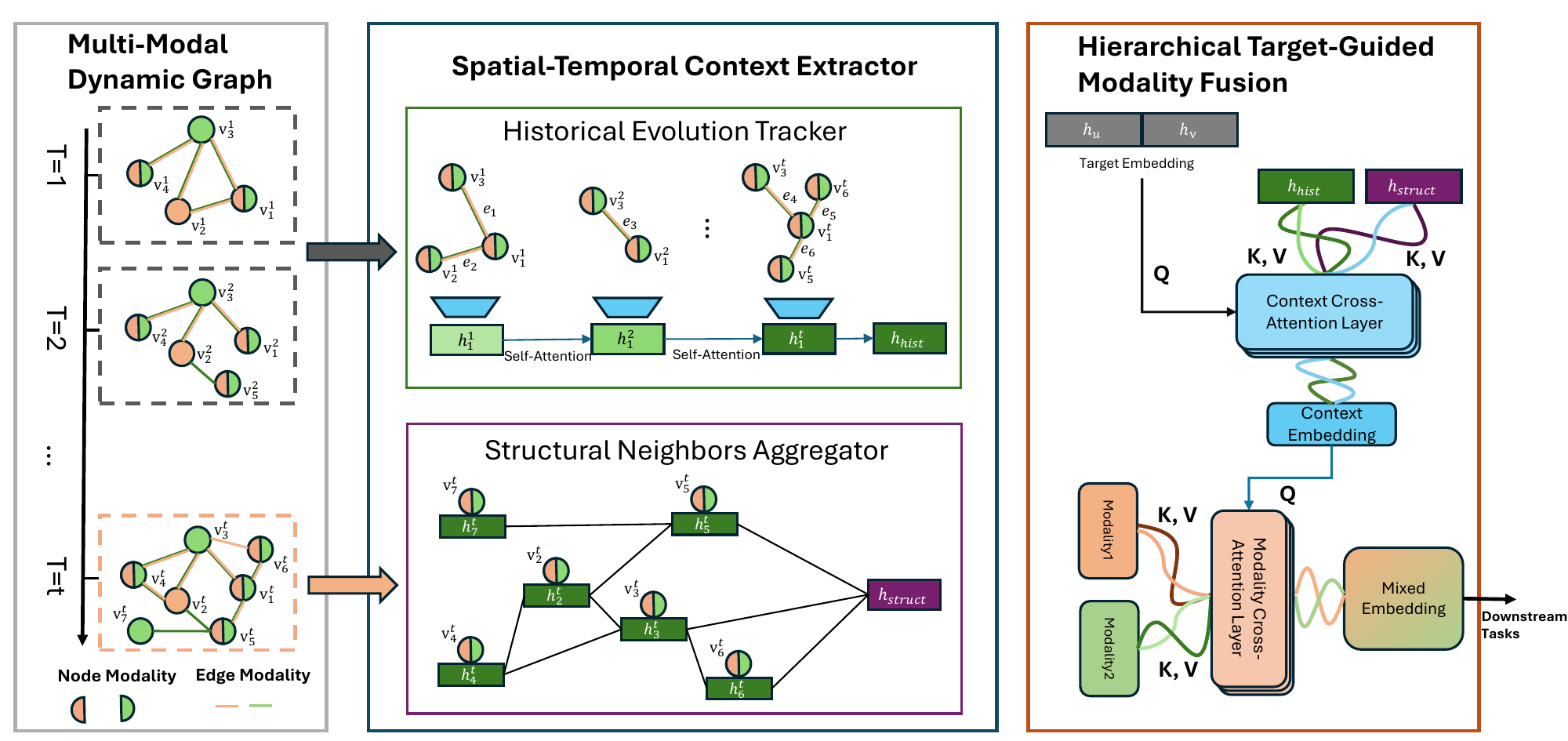}
    \caption{Overview of the proposed TST-MF architecture. Given a prediction query, the Spatial-Temporal Context Extractor encodes (i) the interaction history with a sequential Transformer to obtain a history embedding, and (ii) the local neighborhood with a graph Transformer to obtain a structural embedding. The Hierarchical Target-Guided Modality Fusion module then performs two-stage cross-attention to perform a context-adaptive fused representation for downstream prediction.}
    \label{fig:framework}
\end{figure}

\section{Target-aware Spatial-Temporal Modality Fusion}

\subsection{Observation and Motivation}

The main results reveal that the effectiveness of multimodal input in dynamic graphs is neither universal nor monotonic. Although multimodal attributes can provide complementary semantics, their utility is not globally fixed and instead varies with the evolving \emph{spatiotemporal context}. In several situations, competitive performance can already be achieved with minimal or even random features, suggesting that temporal structural patterns dominate the signal and naively adding modalities may introduce noise. In contrast, on other datasets and objectives, informative attributes lead to noticeable gains, yet the improvements remain inconsistent across backbones, indicating that simple concatenation fails to capture \emph{when} and \emph{where} each modality becomes useful. These observations motivate a \emph{context-aware fusion} mechanism that conditions multimodal integration on the neighborhood (spatial context) and history (temporal context), rather than applying a static fusion rule.

Beyond dataset- and task-level variation, we further observe strong \emph{target-dependent} heterogeneity.
In dynamic graphs, the prediction target, the particular nodes being queried, determines what evidence is relevant. The same modality may be informative for some targets but distracting for others, even within the same dataset and task. Therefore, effective multimodal learning calls for \emph{target-aware fusion}, where cross-modal interplay is guided by the target representation, enabling adaptive modality selection and combination for each prediction instance.

Motivated by these findings, we propose \textbf{TST-MF}, a target-aware Spatial-Temporal Multimodal Fusion framework for dynamic graphs.

\subsection{Model Architecture}

As shown in Figure \ref{fig:framework}, our framework consists of two main components: a \emph{Spatial-Temporal Context Extractor} and a \emph{Hierarchical Target-Guided Modality Fusion} module.
Given a prediction query regarding a node group $\mathcal{A}$, which is the target, on a dynamic graph $\mathcal{G}_{<t}$, we first extract its temporal and structural context, and then perform target-guided multimodal fusion to get a task-conditioned representation $\mathbf{z}_r(\mathcal{A}, t)$ for $r \in \mathcal{A}$.

\textbf{Spatial-Temporal Context Extractor.}
This component produces two complementary context representations for each query.
(i) A \emph{historical evolution tracker} encodes the target's interaction history as a temporal sequence and captures long-range dependencies using a sequential transformer, yielding a history embedding $H_r = H^{\mathrm{his}}(\mathcal{H}_r(t))$ where $\mathcal{H}_r(t)$ is the history sequence.
(ii) A \emph{structural neighbor aggregator} summarizes the target's local neighborhood via a structural attention aggregator, yielding a structural embedding $S_r = S^{\mathrm{struct}}(\mathcal{N}_r(t))$ where $\mathcal{N}_r(t)$ is the node components in $\mathcal{H}_r(t)$.
Together, these embeddings characterize the nodes from both temporal evolution and spatial structure.

\textbf{Hierarchical Target-Guided Modality Fusion.}
We fuse modalities in a hierarchical, target-conditioned manner.
First, we construct a \emph{target embedding} $q = f_{\mathrm{task}}(\mathcal{A})$ from the raw embeddings of the queried entities.
Using this target embedding as a query, we perform cross-attention over the history and structural embeddings to obtain a unified \emph{context embedding} $c_r = g^{\mathrm{context}}(q; H_r,S_r)$, which captures not only spatiotemporal context but also target-specific intent.
Second, the context embedding is used to query modality-specific features (e.g., text and image representations) through another cross-attention stage, producing context-adaptive fused representation $m_r = h^{\mathrm{modal}} (c_r; \mathcal{A})$.
This two-stage design enables the model to (1) adaptively retrieve relevant spatiotemporal evidence for the specific prediction target and (2) selectively integrate modalities conditioned on that evidence, rather than relying on naive concatenation.

The pseudocode for the proposed method is provided in Appendix \ref{app: pseudocode}.

\subsection{Experiment Results}
To verify the efficacy of our proposed \textbf{TST-MF}, we conduct experiments on the aforementioned \textbf{MM-DyGraph}. 
The main results are reported in Table~\ref{tab: overall_result_avg_precision_link_prediction}, Figure~\ref{fig:edge_classification}, and Table~\ref{tab: overall_result_node_retrieval}. 
Overall, \textbf{TST-MF} achieves strong performance across datasets and tasks, frequently obtaining the best or second-best results among strong dynamic graph baselines. This demonstrates that conditioning multimodal fusion on both the prediction target and its spatiotemporal context yields more robust representations than target-agnostic or naive fusion strategies.

To isolate the contribution of the fusion mechanism from that of the temporal backbone, we further compare \textbf{TST-MF} with three representative multimodal fusion strategies, namely gated fusion, modality attention, and the mixture-of-experts fusion of MoE fusion~\cite{he2025unigraph}, all built on the same DyGFormer backbone with comparable hidden dimensions. Results are shown in Table~\ref{tab:fusion_comparison} and Appendix~\ref{app: all_results} Table~\ref{tab:fusion_comparison_full}.
\textbf{TST-MF} outperforms all three strategies in most datasets and tasks, suggesting that target-guided fusion is more effective than target-agnostic alternatives.

\subsection{Ablation Study}
We conduct ablation studies to disentangle the contribution of each design choice in \textbf{TST-MF}. Specifically, we consider four variants that remove (i) the target guidance in the context cross-attention layer, (ii) the hierarchical architecture in the target-guided modality fusion, and either (iii) the spatial or (iv) historical context branch in the spatiotemporal context extractor. Furthermore, to test whether a modality is equally useful for every prediction, we mask it at test time and measure the change in each instance's loss. Due to page limit, the result and detailed analysis are provided in Appendix \ref{app:ablation}.

\section{Conclusion}

In this work, we study multimodal learning on dynamic graphs and find that leveraging heterogeneous signals (e.g., text and images) under temporal evolution remains underexplored. 
To facilitate systematic evaluation, we introduce \textbf{MM-DyGraph}, a benchmark comprising diverse real-world multimodal dynamic graph datasets with fine-grained temporal graph evolution, and providing three representative tasks covering link prediction, edge classification, and node retrieval. 
Benchmarking representative state-of-the-art dynamic graph methods on MM-DyGraph reveals that naively incorporating multimodal inputs does not consistently improve performance and can even be detrimental, indicating that existing models underutilize cross-modal complementarity. 
Motivated by these findings, we propose \textbf{TST-MF}, a target-aware spatiotemporal multimodal fusion framework that conditions fusion on the queried prediction target together with its neighborhood and historical context, enabling adaptive modality integration over time. 
Extensive experiments demonstrate that \textbf{TST-MF} outperforms strong baselines across datasets and tasks, validating the effectiveness and generality of our approach and establishing a solid foundation for future research on multimodal dynamic graph learning. 

\section*{Acknowledgements}
This work was supported by Beijing National Research Center for Information Science and Technology under Grant No.BNR2026TD03005 and National Natural Science Foundation of China under Grant No.62402264.

\clearpage

\bibliographystyle{unsrtnat}
\bibliography{sample-base}


\clearpage
\appendix

\section{Dataset}
\label{app:dataset}
\subsection{License}
All used codes and datasets are publicly available and permit usage for research purposes under either the MIT License or Apache License 2.0.

\label{app:statistic}
The statistics of our dataset and comparison with previous datasets are summarized in Table \ref{tab: overall stats of datasets}.

\begin{table*}[!htbp]
\centering
\caption{Statistics of existing and our datasets. Uniquely, our datasets provide multimodal attributes for \emph{both} nodes and edges.}
\label{tab: overall stats of datasets}
\resizebox{\linewidth}{!}{
\begin{tabular}{c||c|ccccccc}
\toprule
 & \textbf{Datasets} & \textbf{\# Nodes} & \textbf{\# Edges} & \textbf{\# Timestamps} & \textbf{Node Figures} & \textbf{Edge Figures} & \textbf{Domain} & \textbf{Bipartite} \\
\midrule
\multirow{8}{40pt}{\centering Previous Dynamic Graph} & tgbn-trade & 255 & 468,245 & 32 & \ding{55} & \ding{55} & Trade & \ding{55} \\ 
 & tgbl-wiki & 9,227 & 157,474 & 152,757 & \ding{55} & \ding{55} & Interaction & \ding{55} \\ 
 & tgbl-review & 352,637 & 4,873,540 & 6,865 & \ding{55} & \ding{55} & E-commerce & \ding{51} \\ 
 & MOOC & 7,144 & 411,749 & 345,600 & \ding{55} & \ding{55} & Interaction & \ding{51} \\ 
 & Enron & 42,711 & 797,907 & 1,006 & \ding{55} & \ding{55} & Business & \ding{55} \\ 
 & ICEWS1819 & 31,796 & 1,100,071 & 730 & \ding{55} & \ding{55} & Knowledge graph & \ding{55}\\ 
 & Stack elec & 397,702 & 1,262,225 & 5,224 & \ding{55} & \ding{55} & Multi-round dialogue & \ding{51}\\ 
 & Stack ubuntu & 674,248 & 1,497,006 & 4,972 & \ding{55} & \ding{55} & Multi-round dialogue & \ding{51}\\ 
\midrule \midrule
\multirow{6}{40pt}{\centering Previous Multimodal Graph} & Goodreads & 685,294 & 7,235,084 & - & \ding{51} & \ding{55} & Books & \ding{55} \\ 
 & Ele-fasion & 97,766 & 199,602 & - & \ding{51} & \ding{55} & E-commerce & \ding{55}\\ 
 & Movies & 16,672 & 218,390 & - & \ding{51} & \ding{55} & E-commerce & \ding{55}\\ 
 & Toys & 20,695 & 126,886 & - & \ding{51} & \ding{55} & E-commerce & \ding{55}\\
 & Grocery & 17,074 & 171,340 & - & \ding{51} & \ding{55} & E-commerce & \ding{55}\\
 & Reddit & 99,638 &  1,167,188 & - & \ding{51} & \ding{55} & Business & \ding{55}\\
\midrule \midrule
\multirow{8}{40pt}{\centering Ours} & GDELT & 6,546 & 123,365 & 2,710 & \ding{55} & \ding{51} & Sociopolitics & \ding{55} \\
 & GDELT (full) & 18,918 & 1,458,728 & 32,694 & \ding{55} & \ding{51} & Sociopolitics & \ding{55} \\
 & Google Map City & 353,255 & 377,077 & 928 & \ding{55} & \ding{51} & Point of Interest & \ding{51} \\
 & Google Map City (full) & 1,058,101 & 2,874,839 & 5,856 & \ding{55} & \ding{51} & Point of Interest & \ding{51} \\
 & Amazon - Industrial and Scientific & 365,817 & 555,552 & 4,867 & \ding{51} & \ding{51} & E-commerce & \ding{51} \\
 & Amazon - Office Products & 815,855 & 1,362,819 & 6,575 & \ding{51} & \ding{51} & E-commerce & \ding{51} \\
 & Yelp & 535,362 & 1,169,612 & 730 & \ding{51} & \ding{55} & Business & \ding{51} \\
 & Yelp (full) & 2,138,275 & 6,990,280 & 6,036 & \ding{51} & \ding{55} & Business & \ding{51} \\
\bottomrule
\end{tabular}}
\end{table*}

\subsection{Statistical Analysis}
\begin{figure*}[h]
    \centering
    \caption{Dataset Statistical Analysis.}
    \label{fig:ds_dist_stat}
    \includegraphics[width=0.95\linewidth]{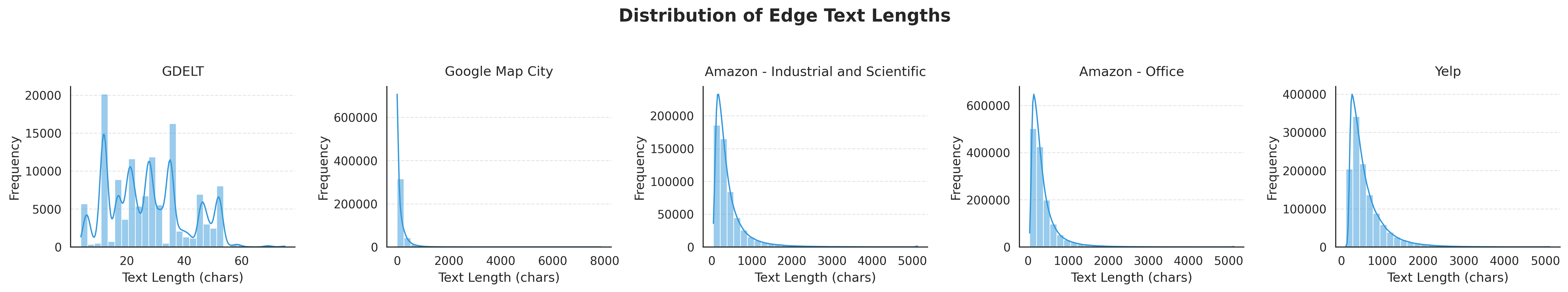}
    \includegraphics[width=0.95\linewidth]{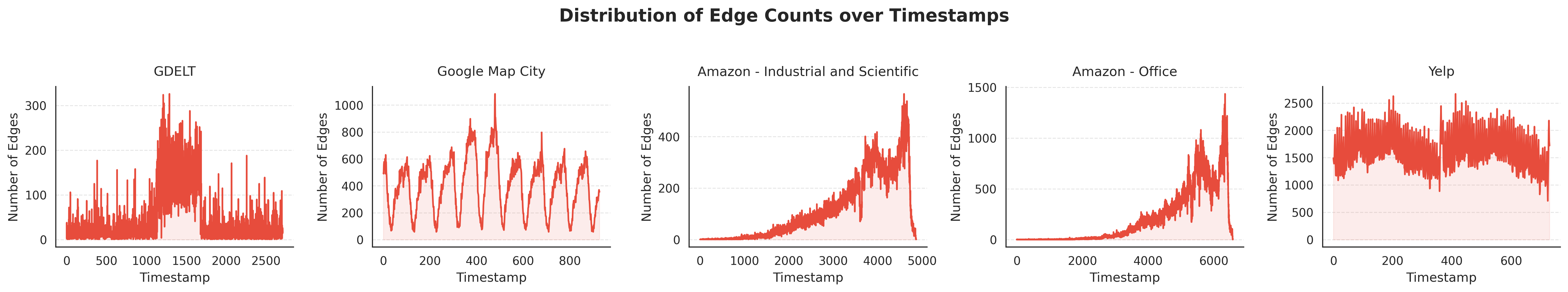}
    \includegraphics[width=0.95\linewidth]{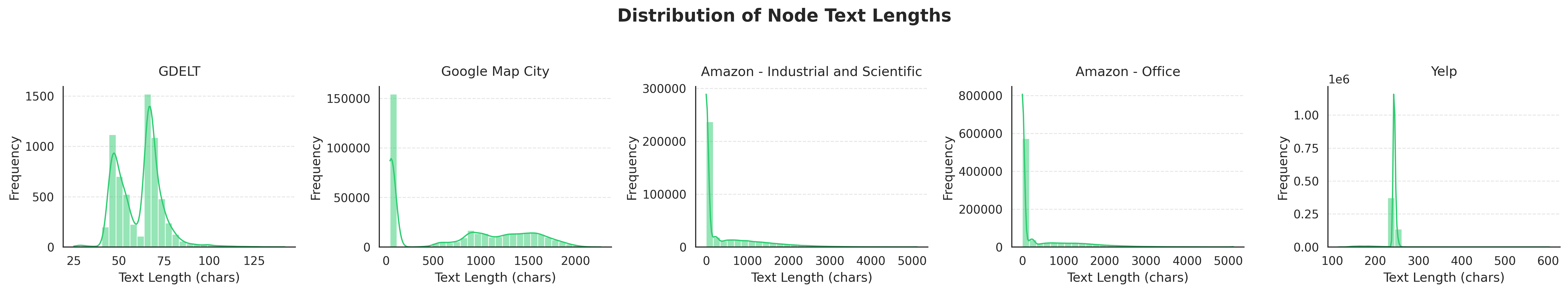}
    \includegraphics[width=0.95\linewidth]{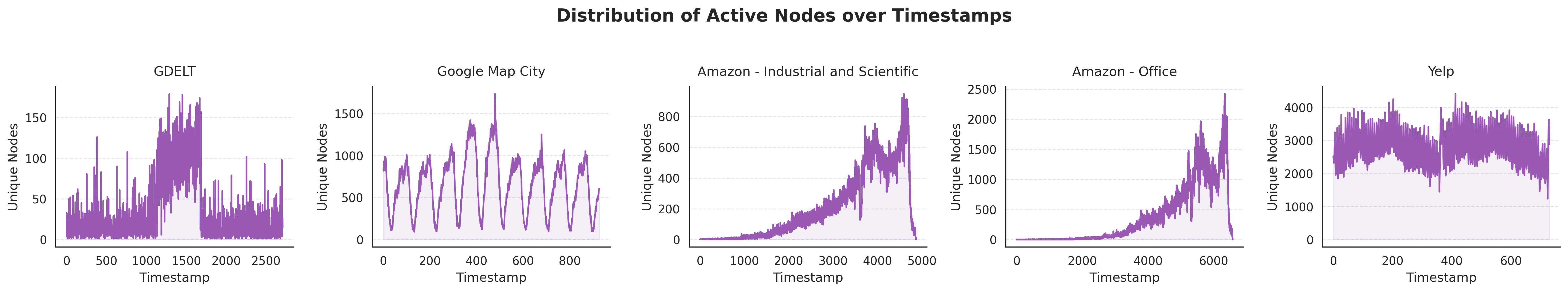}
    \includegraphics[width=0.95\linewidth]{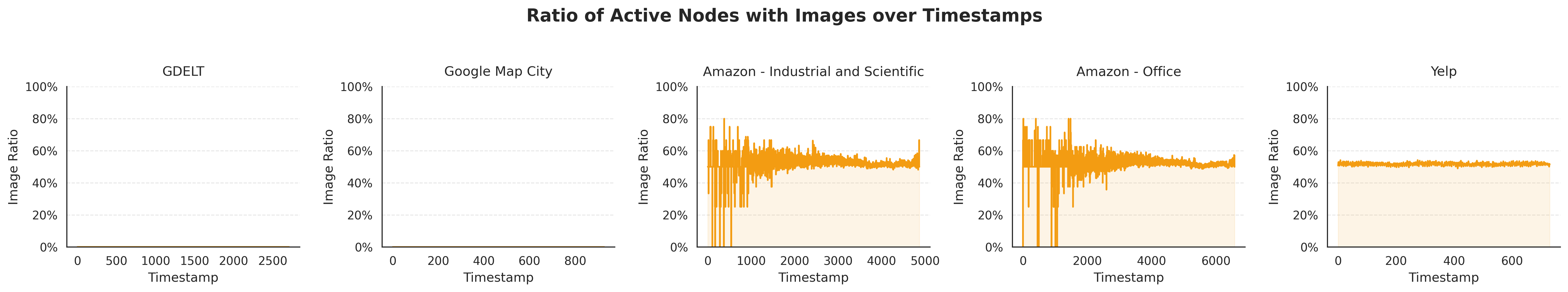}
    \includegraphics[width=0.95\linewidth]{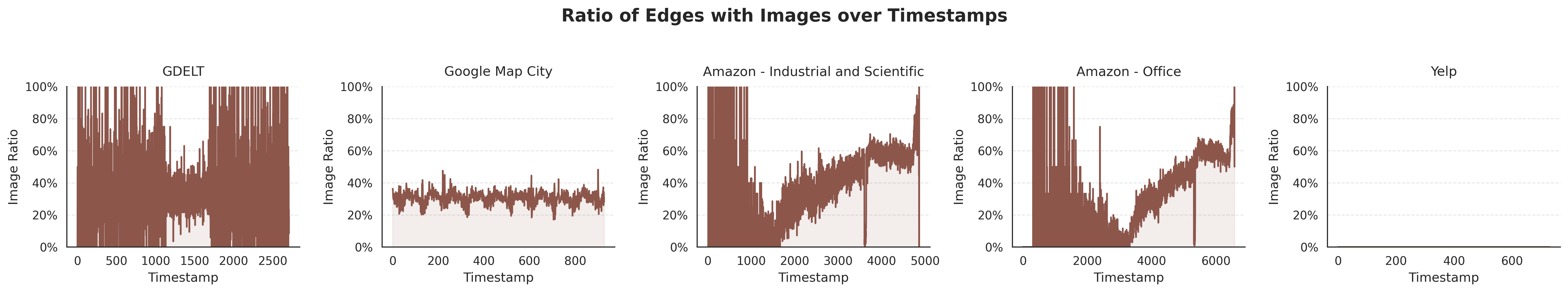}
\end{figure*}

We present a comprehensive statistical analysis of the datasets, examining the distributions of text/image attributes and temporal dynamics in Figure~\ref{fig:ds_dist_stat}.

The distribution of text lengths for edges and nodes reveals distinct domain-specific characteristics. (1) \textbf{Edge Text Lengths}: GDELT presents a multimodal distribution concentrated at shorter lengths, consistent with the heterogeneous endogeneity. Google Map City displays a heavy-tailed distribution, featuring sharp peaks with a sparse tail. Yelp and Amazon exhibit unimodal, right-skewed distributions. (2) \textbf{Node Text Lengths}: GDELT displays a distinct bimodal distribution, indicating two dominant types of node attributes. Meanwhile, the bipartite datasets (Amazon, Yelp, and GMCT) show highly skewed distributions with sharp peaks, indicating homogeneity among text attributes.

The temporal evolution of the datasets, depicted in Figure \ref{fig:ds_dist_stat}.2 and Figure \ref{fig:ds_dist_stat}.4, highlights four highly distinct dynamic patterns. (1) Event-Driven Bursts: Instead of simple stationary volatility, GDELT exhibits a prominent complanate peak in the middle of the timeline, suggesting a surge driven by major real-world exogenous events. (2) Periodic Dynamics: Google Map City demonstrates strong, regular periodicity, reflecting clear cyclical patterns in human mobility or platform usage. (3) Continuous Stability: Unlike typical platform adoption curves, Yelp maintains a relatively stable and uniform interaction volume across the entire observed period. (4) Evolutionary Growth: The Amazon datasets exhibit an evolutionary phase: a "cold start" period followed by exponential growth and subsequent structural shifts. Furthermore, across all datasets, the distribution of active nodes closely tracks the edge distribution, indicating simultaneity between node activation and edge formation.

Furthermore, the temporal evolution of image ratios across active nodes and edges reveals complex structural dynamics. Regarding the node image ratios, we observe distinct stabilization behaviors: while Yelp demonstrates long-term stability with a consistent ratio throughout the timeline, the Amazon datasets initially display significant volatility before converging to a stable equilibrium. Conversely, the edge image ratios exhibit dynamic trends corresponding to the underlying interaction volumes. Notably, GDELT presents a prominent complanate valley in its edge image ratio that inversely mirrors the complanate peak observed in its temporal volume, suggesting a dilution of multimedia content. Google Map City exhibits regular periodic oscillations in its edge image ratio, aligning with its cyclical temporal dynamics. Meanwhile, the Amazon datasets exhibit an initial volatile decline followed by a steady, continuous ascent.

\subsection{Case Study}
\label{app:casestudy}
\begin{figure}[h]
    \centering
    \includegraphics[width=\linewidth]{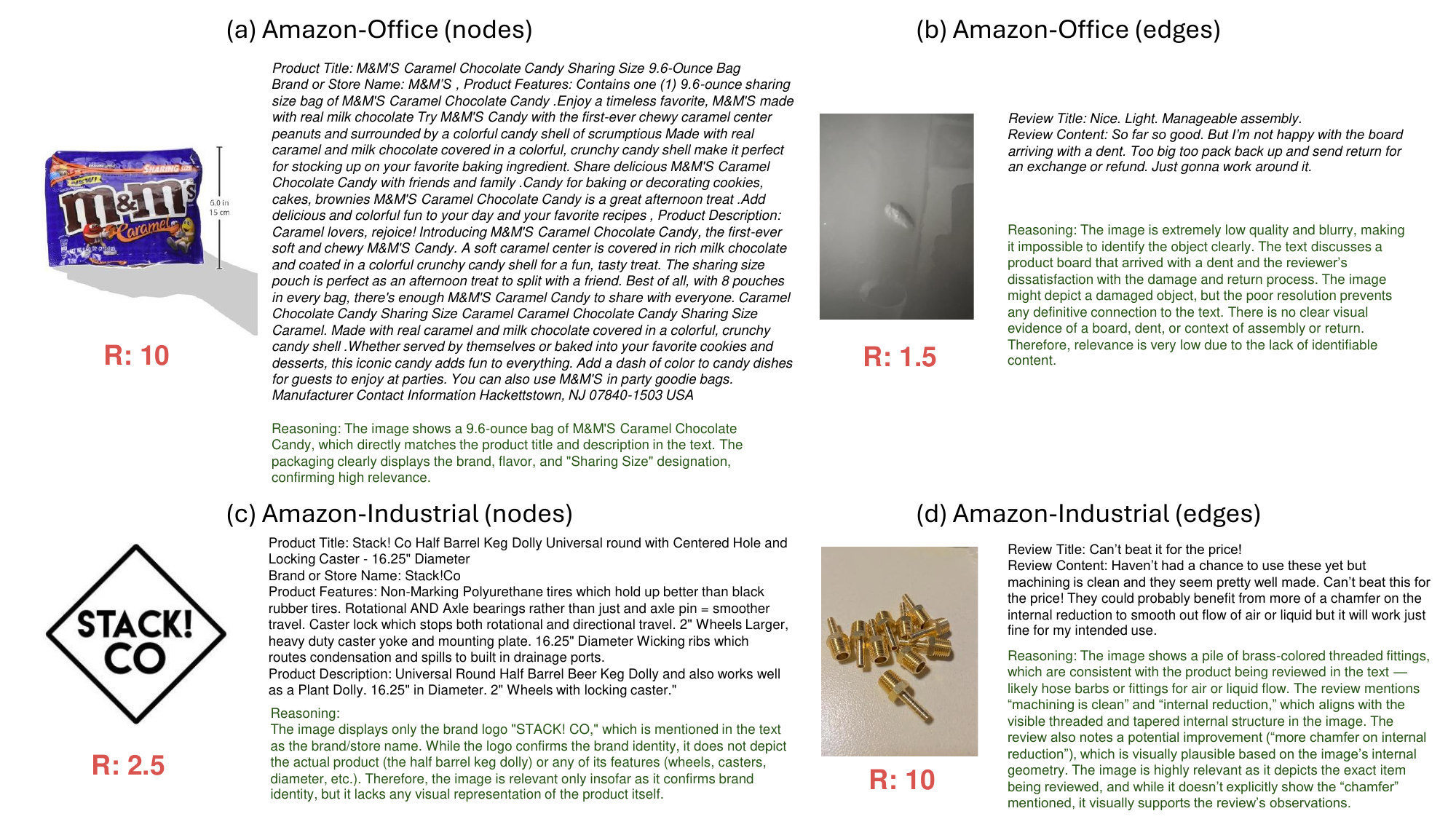}
    \caption{Case studies of VLM-based image-text relevance assessment. The four examples illustrate how the VLM assigns relevance scores to different image-text pairs, including highly matched product images and low-quality visual inputs. These cases show that the VLM can effectively identify semantically aligned image-text pairs, supporting its use as a relevance-based filtering criterion.}
    \label{fig:casestudy}
\end{figure}

To visualize the performance of the VLM-based quality assessment, we demonstrate some case studies on Amazon datasets in Figure \ref{fig:casestudy}. Across the four case studies, the VLM relevance scores are generally consistent with human judgment and effectively reflect the degree of semantic alignment between each image-text pair. In case (a), the image directly depicts the M\&M’S Caramel Chocolate Candy product, and key information such as the brand, flavor, packaging, and sharing-size format is clearly aligned with the text, which justifies the maximum relevance score. In contrast, case (b) receives a very low relevance score because the image is blurry and visually uninformative. The text mentions a damaged board, while the image does not provide sufficient evidence to identify the object or support the review content. Case (c) shows a brand-only image with limited product-level information. Although the image displays the “STACK! CO” logo, which matches the brand name in the text, it does not show the actual half-barrel keg dolly or any of the product attributes described in the text, such as the wheels, locking caster, diameter, or drainage design. Therefore, the VLM assigns a low relevance score, indicating that merely matching the brand identity is insufficient for establishing a strong image-text correspondence. Finally, in case (d), the image clearly shows brass-colored threaded fittings that match the review content about machining quality, internal reduction, and air/liquid flow, resulting in a high relevance score. Overall, these examples suggest that the VLM-based relevance assessment provides reliable judgments. The VLM assigns high scores to image-text pairs with clear semantic alignment while penalizing blurry, uninformative, or only weakly related images. This supports our use of the relevance score as the criterion for filtering image-text pairs.

\section{Pseudocode of TST-MF} \label{app: pseudocode}

\begin{algorithm}[H]
\caption{\textsc{TST-MF}: Task-Guided Spatio-Temporal Multimodal Fusion}
\label{alg:tst-mf}
\begin{algorithmic}[1]
\Require A batch of temporal interactions $\mathcal{B}=\{(u_i,v_i,t_i)\}_{i=1}^{B}$;
historical dynamic graph $\mathcal{G}_{<t}$; node modality features
$\{\mathbf{X}^{(m)}\}_{m=1}^{M}$; edge features $\mathbf{E}$; patch size $p$.
\Ensure Source and destination embeddings $\mathbf{z}_u,\mathbf{z}_v$ for downstream tasks.

\State $\mathbf{q} \gets \mathrm{TaskProj}([\mathbf{x}_{u}\,\|\,\mathbf{x}_{v}])$
\Comment{Construct target-guided task query.}

\For{$r \in \{u,v\}$}
    \Comment{Encode source and destination nodes symmetrically.}
    \State $(\mathcal{N}_r,\mathcal{E}_r,\mathcal{T}_r) \gets
    \mathrm{SampleTemporalNeighbors}(r,t,\mathcal{G}_{<t})$
    \Comment{Retrieve historical context.}

    \State $\mathcal{H}_r \gets
    \mathrm{BuildPatchSequence}(\mathcal{N}_r,\mathcal{E}_r,\mathcal{T}_r,
    \mathbf{X},\mathbf{E},p)$
    \Comment{Patch related signals.}

    \State $\mathbf{H}^{\mathrm{hist}}_r \gets
    \mathrm{HistoricalEvolutionTracker}(\mathcal{H}_r)$
    \Comment{Model temporal evolution by self-attention.}

    \State $\mathbf{S}^{\mathrm{struct}}_r \gets
    \mathrm{StructuralNeighborAggregator}(\mathbf{x}_r,
    \{\mathbf{x}_j:j\in\mathcal{N}_r\})$
    \Comment{Aggregate structural context.}

    \State $\mathbf{c}_r \gets
    \mathrm{ContextCrossAttn}
    (\mathbf{Q}=\mathbf{q},
     \mathbf{K},\mathbf{V}=\{\mathbf{H}^{\mathrm{hist}}_r,
     \mathbf{S}^{\mathrm{struct}}_r\})$
    \Comment{Fuse historical and structural context.}

    \State $\mathbf{m}_r \gets
    \mathrm{ModalityCrossAttn}
    (\mathbf{Q}=\mathbf{c}_r,
     \mathbf{K},\mathbf{V}=\{\mathbf{x}^{(m)}_r\}_{m=1}^{M})$
    \Comment{Select task-relevant modalities.}

    \State $\mathbf{z}_r \gets \mathrm{MLP}(\mathbf{m}_r)$
    \Comment{Project to final node representation.}
\EndFor

\State \Return $\mathbf{z}_u,\mathbf{z}_v$
\Comment{Used by the task-specific prediction head.}
\end{algorithmic}
\end{algorithm}

\section{Evaluation Settings}\label{app:details}
\subsection{Environment}
\label{app:env}

We conduct training with the following hardware configurations:

\begin{itemize}[leftmargin=*, nosep]
\item Operating System: Ubuntu 20.04.6 LTS
\item CPU: AMD EPYC 7763 64-Core Processor
\item GPU: NVIDIA A100-SXM4-40GB GPU and NVIDIA A100-SXM4-80GB GPU
\end{itemize}

\subsection{Hyperparameters}

We involved models DyGFormer, TGAT, GraphMixer, CAWN, FreeDyG, TCL, TPNet, SALoM, Gated Fusion, Modality Attention, MoE fusion, and TST-MF, resulting in
a total of 12 models.
The model architectures and training configurations are summarized in
Table \ref{tab:common_setting} and \ref{tab:model_setting}. For each method, we conducted \emph{three} runs with different random seeds 0, 1, 2. 

\begin{table}[!htbp]
    \centering
    \caption{Common hyperparameters, except that GraphMixer uses a dropout rate of 0.5 on edge classification.}
    \label{tab:common_setting}
    \begin{tabular}{l l}
      \toprule
      \textbf{Hyperparameter} & \textbf{Value} \\
      \midrule
      Batch size & 256 \\
      Optimizer & Adam \\
      Learning rate & $1 \times 10^{-4}$ \\
      Weight decay & 0.0 \\
      Dropout & 0.1 \\
      Max epochs & 200 \\
      Early stopping patience & 5 \\
      Test interval (epochs) & 1 \\
      \#Runs & 3 \\
      Validation ratio & 0.15 \\
      Test ratio & 0.15 \\
      \bottomrule
    \end{tabular}
\end{table}

\begin{table}[!htbp]
    \centering
    \caption{Model-specific hyperparameter. $^{\dagger}$ marks a maximum input sequence length, and SALoM lists both. For CAWN, \#Layers and \#Heads are the walk length and the walk-encoder heads. The fusion module of modality attention adds 4 heads.}
    \label{tab:model_setting}
    \begin{tabular}{l c c c c r}
      \toprule
      \textbf{Model} & \textbf{\#Neighbors} & \textbf{\#Layers} & \textbf{\#Heads} & \textbf{Sampling} & \textbf{\#Params} \\
      \midrule
      TGAT & 20 & 2 & 2 & recent & 50.7M \\
      CAWN & 32 & 1 & 8 & interval & 161.3M \\
      TCL & 20 & 2 & 2 & recent & 9.3M \\
      GraphMixer & 20 & 2 & -- & recent & 7.1M \\
      FreeDyG & 32 & 2 & -- & recent & 57.0M \\
      TPNet & 20 & 2 & -- & recent & 57.4M \\
      SALoM & 10/512$^{\dagger}$ & 1 & 2 & recent & 276.2M \\
      DyGFormer & 48$^{\dagger}$ & 2 & 2 & recent & 6.2M \\
      Gated fusion & 48$^{\dagger}$ & 2 & 2 & recent & 5.9M \\
      Modality attention & 48$^{\dagger}$ & 2 & 2 & recent & 7.1M \\
      MoE fusion & 48$^{\dagger}$ & 2 & 2 & recent & 23.7M \\
      TST-MF & 48$^{\dagger}$ & 2 & 2 & recent & 12.0M \\
      \bottomrule
    \end{tabular}
\end{table}

\clearpage
\subsection{Prompt} \label{app: prompt}

The prompt used in VLM-based quality assessment is shown in Listing \ref{lst:prompt}.

\begin{lstlisting}[language=prompt, caption={Prompt used in VLM-based quality assessment.}, label={lst:prompt}]
### Task
Evaluate the relationship between the provided image and the following text.

### Text
{text}

{extra_image_description (optional)}

### Evaluation Criteria
We want to measure the following aspect:
- Relevance: How related is the image to the text? (e.g., they depict the same item or topic. The image might provide extra information not mentioned in the text, which is completely fine and should not lower the relevance score).

### Output Instructions
Please explain your reasoning in a few sentences first.
After your reasoning, provide two scores, each ranging from 0 to 10 (you can use decimals like 7.3):
- relevance_score
- relevance_confidence (Your confidence in the Relevance Score)

### Output Format
Output your final response STRICTLY in the following format. Do NOT copy the placeholder numbers, make your own judgments. Do not add any text after the JSON block.

Reasoning:
[Your reasoning here]

```json
{{"relevance_score": [0-10 score], "relevance_confidence": [0-10 score]}}
```
\end{lstlisting}

\newpage
\section{Supplementary Empirical Results} \label{app: all_results}

The AUC-ROC results of different methods on the link prediction task are provided in Table \ref{tab: overall_result_auc_link_prediction}. The Precision, Recall and F-score results on the edge classification task are provided in Table \ref{tab: overall_result_avg_precision_edge_classification}, Table \ref{tab: overall_result_avg_recall_edge_classification} and Table \ref{tab: overall_result_avg_fscore_edge_classification}. The Hits@1 and Hits@3 results are provided in Table \ref{tab: overall_result_node_retrieval_h1} and Table \ref{tab: overall_result_node_retrieval_h3}, respectively. The comparison of multimodal fusion strategies on all metrics is reported in Table~\ref{tab:fusion_comparison_full}.

Our pruning process increases multimodal information density, but also changes the retained graph size and modality availability. To assess the sensitivity of our findings to this preprocessing, we additionally evaluate models on the complete GDELT and GMCT datasets, preserving the interactions available before pruning. The results are provided in Tables~\ref{tab:supp_complete_lp}--\ref{tab:supp_complete_nr}. Results on the complete Yelp dataset are omitted because its scale makes evaluation infeasible for most evaluated models under single NVIDIA-A100 GPU setup. The results show that our main observations remain broadly consistent after preserving the original modality sparsity.

Moreover, we examine how different missing modality processing choices affects the benchmark results, we compare four imputation strategies: (i) zero filling, which replaces a missing embedding with a zero vector; (ii) mean filling, which uses the mean of the available embeddings; (iii) learned embedding, which uses a trainable vector to represent the missing modality; and (iv) duplication, which uses the corresponding other modality embedding. The results are reported in Tables~\ref{tab:supp_missing_lp_ap}--\ref{tab:supp_missing_nr_h10}. This comparison examines both the sensitivity of absolute performance to imputation and the stability of the relative performance of different methods. Results indicate that the different imputation strategies yield comparable performance, and the relative performance trends remain consistent.

What's more, our default node-retrieval protocol ranks the true destination against 100 randomly sampled negative destinations. Although this provides a consistent evaluation setting, random negatives can be easy to distinguish on some datasets, leading to near-saturated Hits@10 scores. We therefore additionally evaluate historical negative sampling, which constructs negative candidates from nodes that have recently interacted with the query source before the query time, excluding the current positive destination. This setting emphasizes historically relevant alternatives and provides a more challenging ranking evaluation. And to distinguish the retrieval of previously observed connections from the discovery of new ones, we further report recurrence-aware results. A test interaction is classified as repeated if its source--destination pair has appeared before the query time, and as novel otherwise. Table~\ref{tab:supp_recurrence_h10} reports overall Hits@10 across all test interactions, while Tables~\ref{tab:supp_recurrence_repeated_h10} and~\ref{tab:supp_recurrence_novel_h10} report results on repeated and novel interactions, respectively. Under the stricter protocol, the the metric no longer approaches saturation, confirming that the historical negative sampling is substantially more discriminative.

\begin{table*}[h]
\centering
\small
\caption{AUC-ROC (\%) for link prediction. “\textit{tr.}” and “\textit{in.}” refer to the transductive and inductive setting, respectively. The best results are highlighted in bold and the second-best results are underlined.}
\label{tab: overall_result_auc_link_prediction}
\resizebox{\linewidth}{!}{
\begin{tabular}{c|c|c|ccccccccc}
\toprule
 & \textbf{Datasets} & \textbf{Feature} & \textbf{TGAT} & \textbf{CAWN} & \textbf{TCL} & \textbf{GraphMixer} & \textbf{FreeDyG} & \textbf{TPNet} & \textbf{SALoM} & \textbf{DyGFormer} & \textbf{TST-MF} \\
\midrule
\multirow{15}{*}{\textit{tr.}}

& \multirow{3}{*}{GDELT}
& Random
& 88.30 $\pm$ 1.96 & \underline{92.22 $\pm$ 0.46} & 91.07 $\pm$ 0.03 & 90.70 $\pm$ 0.12 & 90.73 $\pm$ 0.31 & 91.61 $\pm$ 0.54 & 90.26 $\pm$ 1.93 & 91.56 $\pm$ 0.08 & \textbf{92.23 $\pm$ 0.25} \\
& & Text
& 88.82 $\pm$ 1.21 & \underline{92.60 $\pm$ 0.08} & 91.08 $\pm$ 0.22 & 91.36 $\pm$ 0.35 & 91.54 $\pm$ 0.34 & 91.76 $\pm$ 0.21 & 90.92 $\pm$ 0.60 & 92.05 $\pm$ 0.21 & \textbf{93.70 $\pm$ 0.11} \\
& & Text+Image
& 90.16 $\pm$ 0.27 & \underline{92.73 $\pm$ 0.26} & 90.84 $\pm$ 0.32 & 91.15 $\pm$ 0.14 & 91.47 $\pm$ 0.18 & 92.22 $\pm$ 0.22 & 91.18 $\pm$ 0.15 & 91.93 $\pm$ 0.24 & \textbf{93.71 $\pm$ 0.08} \\
\cmidrule{2-12}

& \multirow{3}{*}{GMCT}
& Random
& 73.63 $\pm$ 2.25 & 76.69 $\pm$ 0.02 & 76.54 $\pm$ 0.14 & 76.72 $\pm$ 0.01 & 76.77 $\pm$ 0.01 & 78.46 $\pm$ 0.61 & 61.68 $\pm$ 2.90 & \textbf{94.36 $\pm$ 0.54} & \underline{81.17 $\pm$ 4.79} \\
& & Text
& 81.96 $\pm$ 3.60 & 79.40 $\pm$ 0.15 & 77.34 $\pm$ 0.85 & 79.28 $\pm$ 0.35 & 78.11 $\pm$ 0.31 & 79.08 $\pm$ 0.46 & 58.75 $\pm$ 2.88 & \textbf{95.35 $\pm$ 0.66} & \underline{91.89 $\pm$ 0.76} \\
& & Text+Image
& 81.60 $\pm$ 2.33 & 79.56 $\pm$ 0.05 & 77.42 $\pm$ 0.60 & 79.48 $\pm$ 0.66 & 78.29 $\pm$ 0.25 & 78.78 $\pm$ 0.19 & 57.47 $\pm$ 6.18 & \textbf{95.08 $\pm$ 0.27} & \underline{92.49 $\pm$ 0.66} \\
\cmidrule{2-12}

& \multirow{3}{*}{Amazon - Industrial}
& Random
& 91.92 $\pm$ 0.25 & 91.80 $\pm$ 0.17 & 91.79 $\pm$ 0.25 & \textbf{92.49 $\pm$ 0.18} & 91.69 $\pm$ 0.30 & 92.40 $\pm$ 0.33 & 82.33 $\pm$ 3.29 & 89.27 $\pm$ 1.11 & \underline{92.48 $\pm$ 0.05} \\
\bottomrule
\end{tabular}}
\end{table*}

\begin{table*}[h]
\centering
\small
\caption{Precision (\%) for edge classification. The best results are highlighted in bold and the second-best results are underlined.}
\label{tab: overall_result_avg_precision_edge_classification}
\resizebox{\linewidth}{!}{
\begin{tabular}{c|c|ccccccccc}
\toprule
\textbf{Datasets} & \textbf{Feature} & \textbf{TGAT} & \textbf{CAWN} & \textbf{TCL} & \textbf{GraphMixer} & \textbf{FreeDyG} & \textbf{TPNet} & \textbf{SALoM} & \textbf{DyGFormer} & \textbf{TST-MF} \\
\midrule
\multirow{3}{*}{GDELT}
& Random
& 48.22 $\pm$ 0.33 & \textbf{50.02 $\pm$ 3.28} & 42.98 $\pm$ 3.83 & 48.67 $\pm$ 1.10 & 45.38 $\pm$ 1.53 & 47.34 $\pm$ 2.56 & 40.69 $\pm$ 0.00 & 47.59 $\pm$ 0.17 & \underline{49.23 $\pm$ 1.43} \\
& Text
& 47.96 $\pm$ 0.39 & \underline{49.62 $\pm$ 2.52} & 40.77 $\pm$ 0.00 & 48.22 $\pm$ 0.15 & 44.34 $\pm$ 3.19 & 47.29 $\pm$ 1.25 & 46.33 $\pm$ 1.78 & 47.62 $\pm$ 0.06 & \textbf{52.56 $\pm$ 1.28} \\
& Text+Image
& 48.15 $\pm$ 0.09 & \underline{51.62 $\pm$ 5.99} & 40.77 $\pm$ 0.00 & 48.08 $\pm$ 0.31 & 46.60 $\pm$ 0.21 & 48.89 $\pm$ 2.63 & 47.78 $\pm$ 0.17 & 47.54 $\pm$ 0.10 & \textbf{53.56 $\pm$ 1.70} \\
\cmidrule{1-11}

\multirow{3}{*}{GMCT}
& Random
& 55.26 $\pm$ 0.56 & 55.42 $\pm$ 0.02 & 54.36 $\pm$ 0.20 & \underline{58.09 $\pm$ 0.10} & 50.46 $\pm$ 1.80 & 50.34 $\pm$ 0.66 & 45.13 $\pm$ 0.00 & \textbf{59.44 $\pm$ 0.13} & 51.60 $\pm$ 0.31 \\
& Text
& 54.68 $\pm$ 0.04 & 56.46 $\pm$ 1.92 & 54.56 $\pm$ 0.20 & \underline{58.81 $\pm$ 1.13} & 47.12 $\pm$ 3.45 & 52.66 $\pm$ 0.31 & 45.13 $\pm$ 0.00 & \textbf{59.32 $\pm$ 0.32} & 47.90 $\pm$ 4.80 \\
& Text+Image
& 55.45 $\pm$ 0.29 & 56.64 $\pm$ 2.07 & 54.40 $\pm$ 0.19 & \textbf{58.58 $\pm$ 0.76} & 47.88 $\pm$ 4.75 & 49.49 $\pm$ 3.77 & 47.71 $\pm$ 4.46 & \underline{56.77 $\pm$ 1.72} & 53.45 $\pm$ 1.57 \\
\cmidrule{1-11}

\multirow{3}{*}{Amazon - Industrial}
& Random
& 48.05 $\pm$ 0.87 & 53.14 $\pm$ 3.78 & 47.27 $\pm$ 0.28 & \textbf{57.29 $\pm$ 1.93} & 46.05 $\pm$ 0.37 & 45.95 $\pm$ 0.63 & 42.97 $\pm$ 2.16 & \underline{56.46 $\pm$ 2.26} & 47.96 $\pm$ 0.01 \\
\bottomrule
\end{tabular}}
\end{table*}

\begin{table*}[h]
\centering
\small
\caption{Recall (\%) for edge classification. The best results are highlighted in bold and the second-best results are underlined.}
\label{tab: overall_result_avg_recall_edge_classification}
\resizebox{\linewidth}{!}{
\begin{tabular}{c|c|ccccccccc}
\toprule
\textbf{Datasets} & \textbf{Feature} & \textbf{TGAT} & \textbf{CAWN} & \textbf{TCL} & \textbf{GraphMixer} & \textbf{FreeDyG} & \textbf{TPNet} & \textbf{SALoM} & \textbf{DyGFormer} & \textbf{TST-MF} \\
\midrule
\multirow{3}{*}{GDELT}
& Random
& 64.07 $\pm$ 0.04 & \underline{64.20 $\pm$ 0.08} & 63.85 $\pm$ 0.00 & \textbf{64.32 $\pm$ 0.22} & 63.32 $\pm$ 0.51 & 62.55 $\pm$ 1.04 & 63.79 $\pm$ 0.00 & 63.91 $\pm$ 0.15 & 62.50 $\pm$ 1.35 \\
& Text
& 63.98 $\pm$ 0.08 & \underline{64.23 $\pm$ 0.12} & 63.85 $\pm$ 0.00 & \textbf{64.25 $\pm$ 0.05} & 63.58 $\pm$ 0.27 & 63.82 $\pm$ 0.03 & 63.88 $\pm$ 0.11 & 63.94 $\pm$ 0.09 & 63.70 $\pm$ 0.04 \\
& Text+Image
& 64.05 $\pm$ 0.03 & \underline{64.24 $\pm$ 0.12} & 63.85 $\pm$ 0.00 & \textbf{64.26 $\pm$ 0.11} & 63.36 $\pm$ 0.60 & 63.30 $\pm$ 0.88 & 63.96 $\pm$ 0.09 & 63.93 $\pm$ 0.08 & 63.77 $\pm$ 0.07 \\
\cmidrule{1-11}

\multirow{3}{*}{GMCT}
& Random
& 67.10 $\pm$ 0.39 & \underline{67.45 $\pm$ 0.09} & 67.01 $\pm$ 0.27 & 67.36 $\pm$ 0.31 & 66.91 $\pm$ 0.22 & 61.93 $\pm$ 3.65 & 67.18 $\pm$ 0.00 & \textbf{67.86 $\pm$ 0.03} & 53.66 $\pm$ 1.57 \\
& Text
& 67.16 $\pm$ 0.06 & \textbf{67.60 $\pm$ 0.30} & 66.39 $\pm$ 0.92 & \underline{67.46 $\pm$ 0.25} & 67.16 $\pm$ 0.03 & 66.39 $\pm$ 0.65 & 67.18 $\pm$ 0.00 & 67.39 $\pm$ 0.21 & 65.65 $\pm$ 2.66 \\
& Text+Image
& 67.36 $\pm$ 0.15 & \textbf{67.60 $\pm$ 0.23} & 66.61 $\pm$ 0.54 & \underline{67.48 $\pm$ 0.35} & 66.64 $\pm$ 0.94 & 66.53 $\pm$ 1.01 & 67.18 $\pm$ 0.00 & 67.47 $\pm$ 0.14 & 67.18 $\pm$ 0.00 \\
\cmidrule{1-11}

\multirow{3}{*}{Amazon - Industrial}
& Random
& 64.48 $\pm$ 0.71 & \underline{65.15 $\pm$ 0.19} & 64.66 $\pm$ 0.38 & 63.96 $\pm$ 0.30 & 61.76 $\pm$ 2.36 & 63.03 $\pm$ 2.26 & 64.59 $\pm$ 0.00 & \textbf{65.36 $\pm$ 0.14} & 58.26 $\pm$ 1.42 \\
\bottomrule
\end{tabular}}
\end{table*}

\begin{table*}[h]
\centering
\small
\caption{F1 (\%) for edge classification. The best results are highlighted in bold and the second-best results are underlined.}
\label{tab: overall_result_avg_fscore_edge_classification}
\resizebox{\linewidth}{!}{
\begin{tabular}{c|c|ccccccccc}
\toprule
\textbf{Datasets} & \textbf{Feature} & \textbf{TGAT} & \textbf{CAWN} & \textbf{TCL} & \textbf{GraphMixer} & \textbf{FreeDyG} & \textbf{TPNet} & \textbf{SALoM} & \textbf{DyGFormer} & \textbf{TST-MF} \\
\midrule
\multirow{3}{*}{GDELT}
& Random
& 50.95 $\pm$ 0.23 & \textbf{53.45 $\pm$ 0.06} & 49.78 $\pm$ 0.02 & 52.54 $\pm$ 0.75 & 49.82 $\pm$ 0.18 & 50.39 $\pm$ 0.40 & 49.69 $\pm$ 0.00 & 52.96 $\pm$ 0.05 & \underline{53.32 $\pm$ 0.04} \\
& Text
& 50.48 $\pm$ 0.35 & \textbf{53.50 $\pm$ 0.06} & 49.77 $\pm$ 0.00 & 52.10 $\pm$ 0.29 & 50.83 $\pm$ 1.01 & 50.11 $\pm$ 0.25 & 50.04 $\pm$ 0.45 & 52.93 $\pm$ 0.12 & \underline{53.38 $\pm$ 0.04} \\
& Text+Image
& 50.81 $\pm$ 0.16 & \underline{53.45 $\pm$ 0.10} & 49.77 $\pm$ 0.00 & 52.66 $\pm$ 0.86 & 51.78 $\pm$ 0.85 & 51.19 $\pm$ 1.39 & 50.72 $\pm$ 0.88 & 52.81 $\pm$ 0.07 & \textbf{53.64 $\pm$ 0.03} \\
\cmidrule{1-11}

\multirow{3}{*}{GMCT}
& Random
& 59.20 $\pm$ 1.08 & \underline{59.35 $\pm$ 0.11} & 56.99 $\pm$ 2.53 & \textbf{59.93 $\pm$ 0.17} & 54.18 $\pm$ 0.22 & 54.48 $\pm$ 0.12 & 53.99 $\pm$ 0.00 & 58.40 $\pm$ 0.06 & 52.12 $\pm$ 0.08 \\
& Text
& 58.24 $\pm$ 0.21 & 59.27 $\pm$ 0.02 & 58.73 $\pm$ 0.45 & \textbf{59.98 $\pm$ 0.16} & 53.99 $\pm$ 0.00 & 55.32 $\pm$ 0.84 & 53.99 $\pm$ 0.00 & \underline{59.68 $\pm$ 0.35} & 54.72 $\pm$ 1.27 \\
& Text+Image
& 59.41 $\pm$ 0.31 & 59.33 $\pm$ 0.05 & 58.48 $\pm$ 0.65 & \underline{59.88 $\pm$ 0.17} & 54.57 $\pm$ 1.00 & 54.83 $\pm$ 1.15 & 54.04 $\pm$ 0.08 & \textbf{59.89 $\pm$ 0.06} & 56.32 $\pm$ 0.21 \\
\cmidrule{1-11}

\multirow{3}{*}{Amazon - Industrial}
& Random
& 54.80 $\pm$ 0.86 & 55.79 $\pm$ 0.34 & 53.41 $\pm$ 0.64 & \textbf{56.58 $\pm$ 0.13} & 50.86 $\pm$ 0.36 & 51.13 $\pm$ 0.08 & 50.70 $\pm$ 0.00 & \underline{55.83 $\pm$ 0.48} & 51.00 $\pm$ 0.04 \\
\bottomrule
\end{tabular}}
\end{table*}

\begin{table*}[h]
\centering
\small
\caption{Hits@1 (\%) for node retrieval. “\textit{tr.}” and “\textit{in.}” refer to the transductive and inductive setting, respectively. The best results are highlighted in bold and the second-best results are underlined.}
\label{tab: overall_result_node_retrieval_h1}
\resizebox{\linewidth}{!}{
\begin{tabular}{c|c|c|ccccccccc}
\toprule
 & \textbf{Datasets} & \textbf{Feature} & \textbf{TGAT} & \textbf{CAWN} & \textbf{TCL} & \textbf{GraphMixer} & \textbf{FreeDyG} & \textbf{TPNet} & \textbf{SALoM} & \textbf{DyGFormer} & \textbf{TST-MF} \\
\midrule
\multirow{15}{*}{\textit{tr.}}

& \multirow{3}{*}{GDELT}
& Random
& 28.12 $\pm$ 5.67 & \underline{34.88 $\pm$ 0.69} & 31.63 $\pm$ 0.23 & 30.04 $\pm$ 1.52 & 28.29 $\pm$ 0.28 & 33.58 $\pm$ 2.69 & 32.54 $\pm$ 1.34 & 24.67 $\pm$ 5.01 & \textbf{36.72 $\pm$ 0.29} \\
& & Text
& 28.79 $\pm$ 3.26 & 33.23 $\pm$ 0.06 & 31.70 $\pm$ 0.11 & 31.53 $\pm$ 0.78 & 30.90 $\pm$ 1.44 & 33.56 $\pm$ 0.38 & 32.25 $\pm$ 1.04 & \underline{33.99 $\pm$ 3.87} & \textbf{38.25 $\pm$ 0.56} \\
& & Text+Image
& 30.42 $\pm$ 1.64 & 33.65 $\pm$ 1.09 & 31.53 $\pm$ 0.31 & 31.11 $\pm$ 0.64 & 31.04 $\pm$ 0.49 & \underline{36.34 $\pm$ 1.01} & 32.06 $\pm$ 0.77 & 34.19 $\pm$ 4.03 & \textbf{38.31 $\pm$ 0.74} \\
\cmidrule{2-12}

& \multirow{3}{*}{GMCT}
& Random
& 7.45 $\pm$ 2.79 & 11.74 $\pm$ 0.21 & 11.24 $\pm$ 0.47 & 11.09 $\pm$ 0.14 & \underline{34.74 $\pm$ 1.26} & 12.85 $\pm$ 0.06 & 5.46 $\pm$ 0.06 & \textbf{93.35 $\pm$ 3.69} & 31.33 $\pm$ 18.55 \\
& & Text
& 15.04 $\pm$ 3.97 & 11.15 $\pm$ 0.33 & 11.42 $\pm$ 0.13 & 9.09 $\pm$ 2.51 & 35.77 $\pm$ 0.51 & 12.87 $\pm$ 0.42 & 5.50 $\pm$ 0.08 & \textbf{88.78 $\pm$ 4.31} & \underline{70.15 $\pm$ 7.03} \\
& & Text+Image
& 13.63 $\pm$ 2.32 & 11.64 $\pm$ 0.20 & 11.58 $\pm$ 0.40 & 10.51 $\pm$ 1.45 & 35.98 $\pm$ 0.19 & 13.63 $\pm$ 0.27 & 5.30 $\pm$ 0.19 & \textbf{93.29 $\pm$ 1.56} & \underline{73.27 $\pm$ 3.06} \\
\cmidrule{2-12}

& \multirow{3}{*}{Amazon - Industrial}
& Random
& 28.28 $\pm$ 3.48 & 27.74 $\pm$ 1.22 & 27.37 $\pm$ 4.13 & 32.82 $\pm$ 0.07 & 34.31 $\pm$ 2.70 & 30.65 $\pm$ 3.59 & 14.21 $\pm$ 3.75 & \textbf{44.61 $\pm$ 18.94} & \underline{37.99 $\pm$ 3.49} \\
\bottomrule
\end{tabular}}
\end{table*}

\begin{table*}[h]
\centering
\small
\caption{Hits@3 (\%) for node retrieval. “\textit{tr.}” and “\textit{in.}” refer to the transductive and inductive setting, respectively. The best results are highlighted in bold, and the second-best results are underlined.}
\label{tab: overall_result_node_retrieval_h3}
\resizebox{\linewidth}{!}{
\begin{tabular}{c|c|c|ccccccccc}
\toprule
 & \textbf{Datasets} & \textbf{Feature} & \textbf{TGAT} & \textbf{CAWN} & \textbf{TCL} & \textbf{GraphMixer} & \textbf{FreeDyG} & \textbf{TPNet} & \textbf{SALoM} & \textbf{DyGFormer} & \textbf{TST-MF} \\
\midrule
\multirow{15}{*}{\textit{tr.}}

& \multirow{3}{*}{GDELT}
& Random
& 47.26 $\pm$ 6.65 & \underline{55.88 $\pm$ 0.56} & 53.42 $\pm$ 0.19 & 48.30 $\pm$ 3.94 & 46.34 $\pm$ 1.30 & 51.59 $\pm$ 2.51 & 55.21 $\pm$ 2.00 & 35.27 $\pm$ 6.80 & \textbf{57.46 $\pm$ 0.46} \\
& & Text
& 49.37 $\pm$ 4.36 & \underline{55.20 $\pm$ 0.54} & 53.64 $\pm$ 0.14 & 52.96 $\pm$ 1.39 & 53.50 $\pm$ 1.01 & 54.73 $\pm$ 0.82 & 55.01 $\pm$ 1.21 & 51.40 $\pm$ 6.58 & \textbf{58.98 $\pm$ 1.08} \\
& & Text+Image
& 51.56 $\pm$ 2.13 & 55.65 $\pm$ 1.06 & 53.46 $\pm$ 0.21 & 51.71 $\pm$ 2.01 & 51.67 $\pm$ 0.63 & 54.84 $\pm$ 2.20 & \underline{56.03 $\pm$ 0.24} & 52.06 $\pm$ 6.41 & \textbf{59.05 $\pm$ 0.36} \\
\cmidrule{2-12}

& \multirow{3}{*}{GMCT}
& Random
& 20.00 $\pm$ 5.05 & 25.25 $\pm$ 0.31 & 25.20 $\pm$ 0.35 & 25.57 $\pm$ 0.35 & 39.36 $\pm$ 0.84 & 26.29 $\pm$ 0.07 & 11.35 $\pm$ 0.13 & \textbf{100.00 $\pm$ 0.01} & \underline{59.31 $\pm$ 31.43} \\
& & Text
& 30.51 $\pm$ 6.38 & 25.00 $\pm$ 0.18 & 24.90 $\pm$ 0.29 & 23.08 $\pm$ 2.21 & 38.90 $\pm$ 0.37 & 26.56 $\pm$ 0.16 & 11.60 $\pm$ 0.16 & \textbf{100.00 $\pm$ 0.01} & \underline{99.63 $\pm$ 0.34} \\
& & Text+Image
& 29.04 $\pm$ 4.09 & 25.53 $\pm$ 0.18 & 25.24 $\pm$ 0.17 & 25.07 $\pm$ 1.56 & 38.84 $\pm$ 0.27 & 27.28 $\pm$ 0.17 & 11.31 $\pm$ 0.22 & \textbf{100.00 $\pm$ 0.01} & \underline{99.69 $\pm$ 0.69} \\
\cmidrule{2-12}

& \multirow{3}{*}{Amazon - Industrial}
& Random
& 44.03 $\pm$ 4.84 & 44.17 $\pm$ 0.85 & 46.34 $\pm$ 4.80 & 50.38 $\pm$ 0.84 & 46.78 $\pm$ 3.58 & 44.94 $\pm$ 4.79 & 31.47 $\pm$ 4.48 & \textbf{80.80 $\pm$ 21.52} & \underline{63.54 $\pm$ 8.20} \\
\bottomrule
\end{tabular}}
\end{table*}

\begin{table*}[h]
\centering
\small
\caption{Comparison of multimodal fusion strategies on all metrics under the Text+Image setting (\%). All methods share the DyGFormer backbone, and DyGFormer denotes naive feature concatenation. “\textit{tr.}” and “\textit{in.}” refer to the transductive and inductive setting, respectively. The best results are highlighted in bold, and the second-best results are underlined.}
\label{tab:fusion_comparison_full}
\resizebox{\linewidth}{!}{
\begin{tabular}{c|c|c|c|c|ccc|c}
\toprule
\textbf{Task} & \textbf{Metric} & & \textbf{Datasets} & \textbf{DyGFormer} & \textbf{Gated fusion} & \textbf{Modality attention} & \textbf{MoE fusion} & \textbf{TST-MF} \\
\midrule
\multirow{20}{*}{LP} & \multirow{10}{*}{AP} & \multirow{5}{*}{\textit{tr.}} & GDELT & 92.46 $\pm$ 0.26 & 92.74 $\pm$ 0.28 & \underline{93.08 $\pm$ 0.10} & 92.92 $\pm$ 0.26 & \textbf{93.98 $\pm$ 0.05} \\
 &  &  & GMCT & 95.54 $\pm$ 0.10 & \underline{96.41 $\pm$ 0.17} & 96.30 $\pm$ 0.20 & \textbf{96.59 $\pm$ 0.16} & 89.08 $\pm$ 0.72 \\
 &  &  & Amazon - Industrial & 92.56 $\pm$ 0.24 & \underline{92.63 $\pm$ 0.26} & 92.61 $\pm$ 0.61 & 90.89 $\pm$ 2.25 & \textbf{94.10 $\pm$ 0.12} \\
 &  &  & Amazon - Office & 89.00 $\pm$ 0.23 & 90.26 $\pm$ 1.41 & \underline{90.73 $\pm$ 1.02} & 89.59 $\pm$ 0.94 & \textbf{92.67 $\pm$ 0.10} \\
 &  &  & Yelp & 88.55 $\pm$ 5.51 & 91.25 $\pm$ 0.38 & \underline{92.29 $\pm$ 0.17} & 88.56 $\pm$ 5.75 & \textbf{92.65 $\pm$ 0.04} \\
\cmidrule{3-9}
 &  & \multirow{5}{*}{\textit{in.}} & GDELT & 83.05 $\pm$ 0.28 & 83.29 $\pm$ 0.34 & 84.07 $\pm$ 0.31 & \underline{84.14 $\pm$ 0.35} & \textbf{85.44 $\pm$ 0.14} \\
 &  &  & GMCT & \textbf{83.05 $\pm$ 3.36} & \underline{82.46 $\pm$ 1.83} & 81.92 $\pm$ 2.71 & 82.28 $\pm$ 1.28 & 65.49 $\pm$ 1.00 \\
\bottomrule
\end{tabular}}
\end{table*}

\begin{table*}[t]
\centering
\small
\caption{Link prediction results (\%) on the complete GDELT and GMCT datasets. ``\textit{tr.}'' and ``\textit{in.}'' refer to the transductive and inductive settings, respectively. AP refers to Average Precision. The best results are highlighted in bold and the second-best results are underlined.}
\label{tab:supp_complete_lp}
\resizebox{\linewidth}{!}{%
\begin{tabular}{c|c|c|c|cccccc}
\toprule
 & \textbf{Datasets} & \textbf{Metric} & \textbf{Feature} & \textbf{TGAT} & \textbf{CAWN} & \textbf{TCL} & \textbf{GraphMixer} & \textbf{DyGFormer} & \textbf{TST-MF} \\
\midrule
\multirow{12}{*}{\textit{tr.}} & \multirow{6}{*}{GDELT} & \multirow{3}{*}{AP} & Random & 96.19 $\pm$ 0.27 & 97.04 $\pm$ 0.04 & 96.20 $\pm$ 0.11 & 96.06 $\pm$ 0.20 & \underline{97.25 $\pm$ 0.03} & \textbf{97.36 $\pm$ 0.03} \\
 &  &  & Text & 97.40 $\pm$ 0.07 & 97.37 $\pm$ 0.13 & 96.09 $\pm$ 0.15 & 96.96 $\pm$ 0.11 & \underline{97.71 $\pm$ 0.02} & \textbf{97.87 $\pm$ 0.02} \\
 &  &  & Text+Image & 97.42 $\pm$ 0.07 & 97.27 $\pm$ 0.28 & 96.19 $\pm$ 0.25 & 97.03 $\pm$ 0.08 & \underline{97.62 $\pm$ 0.09} & \textbf{97.85 $\pm$ 0.04} \\
\cmidrule{3-10}
 &  & \multirow{3}{*}{AUC-ROC} & Random & 95.97 $\pm$ 0.26 & 96.91 $\pm$ 0.04 & 96.02 $\pm$ 0.15 & 95.94 $\pm$ 0.14 & \underline{97.01 $\pm$ 0.07} & \textbf{97.18 $\pm$ 0.03} \\
 &  &  & Text & 97.30 $\pm$ 0.07 & 97.31 $\pm$ 0.13 & 95.93 $\pm$ 0.18 & 96.87 $\pm$ 0.15 & \underline{97.63 $\pm$ 0.03} & \textbf{97.80 $\pm$ 0.02} \\
 &  &  & Text+Image & 97.33 $\pm$ 0.08 & 97.20 $\pm$ 0.29 & 96.06 $\pm$ 0.28 & 96.97 $\pm$ 0.08 & \underline{97.53 $\pm$ 0.11} & \textbf{97.78 $\pm$ 0.06} \\
\cmidrule{2-10}
 & \multirow{6}{*}{GMCT} & \multirow{3}{*}{AP} & Random & 72.24 $\pm$ 0.43 & 74.92 $\pm$ 0.07 & 72.11 $\pm$ 3.38 & 74.96 $\pm$ 0.01 & \textbf{78.65 $\pm$ 0.05} & \underline{76.11 $\pm$ 0.55} \\
\bottomrule
\end{tabular}
}
\end{table*}

\begin{table*}[t]
\centering
\small
\caption{Edge classification results (\%) on the complete GDELT and GMCT datasets. The best results are highlighted in bold and the second-best results are underlined.}
\label{tab:supp_complete_ec}
\resizebox{\linewidth}{!}{
\begin{tabular}{c|c|c|cccccc}
\toprule
\textbf{Datasets} & \textbf{Metric} & \textbf{Feature} & \textbf{TGAT} & \textbf{CAWN} & \textbf{TCL} & \textbf{GraphMixer} & \textbf{DyGFormer} & \textbf{TST-MF} \\
\midrule
\multirow{9}{*}{GDELT} & \multirow{3}{*}{Precision} & Random & 60.92 $\pm$ 2.70 & \textbf{61.85 $\pm$ 0.58} & 46.09 $\pm$ 4.48 & 60.68 $\pm$ 0.94 & 57.56 $\pm$ 0.26 & \underline{61.23 $\pm$ 0.14} \\
 &  & Text & 49.57 $\pm$ 0.25 & \textbf{63.85 $\pm$ 1.26} & 45.92 $\pm$ 4.34 & 58.91 $\pm$ 10.07 & 60.51 $\pm$ 3.81 & \underline{61.79 $\pm$ 0.67} \\
 &  & Text+Image & 49.43 $\pm$ 0.13 & \textbf{62.49 $\pm$ 1.92} & 43.50 $\pm$ 4.47 & 55.61 $\pm$ 5.35 & 59.74 $\pm$ 4.03 & \underline{61.13 $\pm$ 2.25} \\
\cmidrule{2-9}
 & \multirow{3}{*}{Recall} & Random & 64.69 $\pm$ 0.03 & \underline{65.01 $\pm$ 0.02} & 64.43 $\pm$ 0.40 & \textbf{65.09 $\pm$ 0.06} & \textbf{65.09 $\pm$ 0.02} & 64.18 $\pm$ 0.34 \\
 &  & Text & 64.39 $\pm$ 0.35 & 65.04 $\pm$ 0.04 & 64.27 $\pm$ 0.35 & 65.00 $\pm$ 0.10 & \underline{65.20 $\pm$ 0.09} & \textbf{65.41 $\pm$ 0.09} \\
 &  & Text+Image & 64.50 $\pm$ 0.21 & 64.94 $\pm$ 0.02 & 64.20 $\pm$ 0.39 & 64.97 $\pm$ 0.18 & \underline{65.15 $\pm$ 0.22} & \textbf{65.48 $\pm$ 0.01} \\
\cmidrule{2-9}
 & \multirow{3}{*}{F1} & Random & 54.73 $\pm$ 0.20 & \underline{55.24 $\pm$ 0.07} & 52.12 $\pm$ 2.00 & \textbf{55.66 $\pm$ 0.02} & 55.17 $\pm$ 0.14 & 54.91 $\pm$ 0.21 \\
\bottomrule
\end{tabular}
}
\end{table*}

\begin{table*}[t]
\centering
\small
\caption{Node retrieval results (\%) on the complete GDELT and GMCT datasets. ``\textit{tr.}'' and ``\textit{in.}'' refer to the transductive and inductive settings, respectively. The best results are highlighted in bold and the second-best results are underlined.}
\label{tab:supp_complete_nr}
\resizebox{\linewidth}{!}{
\begin{tabular}{c|c|c|c|cccccc}
\toprule
 & \textbf{Datasets} & \textbf{Metric} & \textbf{Feature} & \textbf{TGAT} & \textbf{CAWN} & \textbf{TCL} & \textbf{GraphMixer} & \textbf{DyGFormer} & \textbf{TST-MF} \\
\midrule
\multirow{18}{*}{\textit{tr.}} & \multirow{9}{*}{GDELT} & \multirow{3}{*}{Hits@1} & Random & 51.39 $\pm$ 1.29 & \underline{55.65 $\pm$ 0.32} & 50.30 $\pm$ 0.38 & 50.14 $\pm$ 0.95 & 31.33 $\pm$ 7.78 & \textbf{58.59 $\pm$ 0.05} \\
 &  &  & Text & 58.39 $\pm$ 0.65 & 55.33 $\pm$ 1.70 & 50.00 $\pm$ 0.61 & 54.93 $\pm$ 0.39 & \underline{59.63 $\pm$ 0.19} & \textbf{61.35 $\pm$ 0.46} \\
 &  &  & Text+Image & \underline{58.48 $\pm$ 0.19} & 55.66 $\pm$ 2.01 & 49.96 $\pm$ 1.00 & 55.08 $\pm$ 0.20 & 52.54 $\pm$ 7.70 & \textbf{61.18 $\pm$ 0.35} \\
\cmidrule{3-10}
 &  & \multirow{3}{*}{Hits@3} & Random & 73.00 $\pm$ 2.03 & \underline{76.29 $\pm$ 0.13} & 73.47 $\pm$ 0.25 & 70.30 $\pm$ 2.69 & 45.47 $\pm$ 6.98 & \textbf{79.51 $\pm$ 0.11} \\
 &  &  & Text & 79.73 $\pm$ 0.49 & 76.93 $\pm$ 1.73 & 72.49 $\pm$ 0.49 & 76.69 $\pm$ 0.39 & \underline{79.91 $\pm$ 0.34} & \textbf{82.26 $\pm$ 0.25} \\
 &  &  & Text+Image & \underline{79.79 $\pm$ 0.20} & 77.25 $\pm$ 1.64 & 73.37 $\pm$ 0.91 & 76.84 $\pm$ 0.43 & 71.04 $\pm$ 9.37 & \textbf{82.01 $\pm$ 0.28} \\
\cmidrule{3-10}
 &  & \multirow{3}{*}{Hits@10} & Random & 88.47 $\pm$ 0.86 & \underline{91.07 $\pm$ 0.68} & 89.17 $\pm$ 0.10 & 88.23 $\pm$ 0.89 & 70.42 $\pm$ 5.32 & \textbf{93.15 $\pm$ 0.18} \\
\bottomrule
\end{tabular}
}
\end{table*}

\begin{table*}[t]
\centering
\small
\caption{Average Precision (\%) for link prediction under different missing-modality filling strategies. ``\textit{tr.}'' and ``\textit{in.}'' refer to the transductive and inductive settings, respectively. The best results are highlighted in bold and the second-best results are underlined.}
\label{tab:supp_missing_lp_ap}
\resizebox{\linewidth}{!}{
\begin{tabular}{c|c|c|cccccc}
\toprule
 & \textbf{Datasets} & \textbf{Feature} & \textbf{TGAT} & \textbf{CAWN} & \textbf{TCL} & \textbf{GraphMixer} & \textbf{DyGFormer} & \textbf{TST-MF} \\
\midrule
\multirow{20}{*}{\textit{tr.}} & \multirow{4}{*}{GDELT} & Zero Filling & 90.24 $\pm$ 0.64 & \underline{93.08 $\pm$ 0.18} & 91.67 $\pm$ 0.14 & 91.18 $\pm$ 0.32 & 92.33 $\pm$ 0.17 & \textbf{93.93 $\pm$ 0.11} \\
 &  & Mean Filling & 90.32 $\pm$ 0.42 & \underline{92.94 $\pm$ 0.09} & 91.69 $\pm$ 0.05 & 91.27 $\pm$ 0.32 & 92.26 $\pm$ 0.32 & \textbf{94.07 $\pm$ 0.01} \\
 &  & Learned Embedding & 89.96 $\pm$ 0.63 & \underline{92.94 $\pm$ 0.48} & 91.71 $\pm$ 0.20 & 91.39 $\pm$ 0.21 & 92.30 $\pm$ 0.18 & \textbf{94.01 $\pm$ 0.16} \\
 &  & Duplication & 90.76 $\pm$ 0.34 & \underline{93.08 $\pm$ 0.25} & 91.47 $\pm$ 0.19 & 91.38 $\pm$ 0.20 & 92.46 $\pm$ 0.26 & \textbf{93.98 $\pm$ 0.05} \\
\cmidrule{2-9}
 & \multirow{4}{*}{GMCT} & Zero Filling & 82.47 $\pm$ 3.32 & 79.58 $\pm$ 0.08 & 78.63 $\pm$ 0.10 & 79.57 $\pm$ 0.38 & \textbf{95.60 $\pm$ 0.27} & \underline{87.59 $\pm$ 0.99} \\
 &  & Mean Filling & 84.19 $\pm$ 2.11 & 79.59 $\pm$ 0.03 & 78.11 $\pm$ 0.23 & 79.54 $\pm$ 0.22 & \textbf{96.16 $\pm$ 0.32} & \underline{88.55 $\pm$ 1.57} \\
 &  & Learned Embedding & 81.47 $\pm$ 2.19 & 79.32 $\pm$ 0.18 & 78.55 $\pm$ 0.53 & 79.15 $\pm$ 0.13 & \textbf{96.29 $\pm$ 0.41} & \underline{89.26 $\pm$ 1.57} \\
\bottomrule
\end{tabular}
}
\end{table*}

\begin{table*}[t]
\centering
\small
\caption{Precision (\%) for edge classification under different missing-modality filling strategies. The best results are highlighted in bold and the second-best results are underlined.}
\label{tab:supp_missing_ec_precision}
\resizebox{\linewidth}{!}{
\begin{tabular}{c|c|cccccc}
\toprule
\textbf{Datasets} & \textbf{Feature} & \textbf{TGAT} & \textbf{CAWN} & \textbf{TCL} & \textbf{GraphMixer} & \textbf{DyGFormer} & \textbf{TST-MF} \\
\midrule
\multirow{4}{*}{GDELT} & Zero Filling & 47.83 $\pm$ 0.31 & \textbf{51.86 $\pm$ 3.13} & 40.77 $\pm$ 0.00 & 48.19 $\pm$ 0.50 & 47.54 $\pm$ 0.16 & \underline{49.39 $\pm$ 1.96} \\
 & Mean Filling & 47.93 $\pm$ 0.17 & \underline{50.12 $\pm$ 3.53} & 42.44 $\pm$ 2.89 & 47.85 $\pm$ 0.30 & 47.54 $\pm$ 0.08 & \textbf{51.56 $\pm$ 0.47} \\
 & Learned Embedding & 47.96 $\pm$ 0.11 & \textbf{50.77 $\pm$ 2.57} & 44.51 $\pm$ 6.47 & 48.27 $\pm$ 0.23 & 47.46 $\pm$ 0.14 & \underline{50.54 $\pm$ 1.90} \\
 & Duplication & 48.15 $\pm$ 0.09 & \underline{51.62 $\pm$ 5.99} & 40.77 $\pm$ 0.00 & 48.08 $\pm$ 0.31 & 47.54 $\pm$ 0.10 & \textbf{53.56 $\pm$ 1.70} \\
\midrule
\multirow{4}{*}{GMCT} & Zero Filling & 55.39 $\pm$ 0.40 & 56.55 $\pm$ 2.10 & 54.90 $\pm$ 0.51 & \underline{58.51 $\pm$ 0.02} & \textbf{59.22 $\pm$ 0.31} & 53.83 $\pm$ 1.21 \\
 & Mean Filling & 55.16 $\pm$ 0.28 & 55.46 $\pm$ 0.09 & 54.28 $\pm$ 0.28 & \underline{58.35 $\pm$ 0.25} & \textbf{58.56 $\pm$ 2.63} & 52.31 $\pm$ 0.91 \\
 & Learned Embedding & 54.89 $\pm$ 0.14 & 55.43 $\pm$ 0.03 & 54.37 $\pm$ 0.03 & \textbf{58.57 $\pm$ 0.61} & \underline{57.60 $\pm$ 3.35} & 50.79 $\pm$ 4.92 \\
\bottomrule
\end{tabular}
}
\end{table*}

\begin{table*}[t]
\centering
\small
\caption{Recall (\%) for edge classification under different missing-modality filling strategies. The best results are highlighted in bold and the second-best results are underlined.}
\label{tab:supp_missing_ec_recall}
\resizebox{\linewidth}{!}{
\begin{tabular}{c|c|cccccc}
\toprule
\textbf{Datasets} & \textbf{Feature} & \textbf{TGAT} & \textbf{CAWN} & \textbf{TCL} & \textbf{GraphMixer} & \textbf{DyGFormer} & \textbf{TST-MF} \\
\midrule
\multirow{4}{*}{GDELT} & Zero Filling & 64.01 $\pm$ 0.04 & \underline{64.28 $\pm$ 0.09} & 63.85 $\pm$ 0.00 & \textbf{64.29 $\pm$ 0.23} & 63.91 $\pm$ 0.04 & 63.89 $\pm$ 0.64 \\
 & Mean Filling & 63.98 $\pm$ 0.10 & \textbf{64.18 $\pm$ 0.09} & 63.85 $\pm$ 0.01 & \underline{64.16 $\pm$ 0.13} & 63.92 $\pm$ 0.05 & 63.99 $\pm$ 0.16 \\
 & Learned Embedding & 64.01 $\pm$ 0.03 & \underline{64.26 $\pm$ 0.12} & 63.86 $\pm$ 0.01 & \textbf{64.34 $\pm$ 0.09} & 63.88 $\pm$ 0.08 & 64.13 $\pm$ 0.09 \\
 & Duplication & 64.05 $\pm$ 0.03 & \underline{64.24 $\pm$ 0.12} & 63.85 $\pm$ 0.00 & \textbf{64.26 $\pm$ 0.11} & 63.93 $\pm$ 0.08 & 63.77 $\pm$ 0.07 \\
\midrule
\multirow{4}{*}{GMCT} & Zero Filling & 67.17 $\pm$ 0.08 & \textbf{67.52 $\pm$ 0.13} & 66.26 $\pm$ 1.27 & \underline{67.37 $\pm$ 0.35} & \textbf{67.52 $\pm$ 0.17} & 64.27 $\pm$ 3.20 \\
 & Mean Filling & 67.00 $\pm$ 0.27 & 67.45 $\pm$ 0.08 & 66.98 $\pm$ 0.19 & \underline{67.55 $\pm$ 0.08} & \textbf{67.62 $\pm$ 0.11} & 64.59 $\pm$ 1.93 \\
 & Learned Embedding & 66.81 $\pm$ 0.47 & \textbf{67.54 $\pm$ 0.07} & 66.50 $\pm$ 0.81 & \underline{67.46 $\pm$ 0.19} & 67.35 $\pm$ 0.18 & 64.92 $\pm$ 1.97 \\
\bottomrule
\end{tabular}
}
\end{table*}

\begin{table*}[t]
\centering
\small
\caption{F1 (\%) for edge classification under different missing-modality filling strategies. The best results are highlighted in bold and the second-best results are underlined.}
\label{tab:supp_missing_ec_f1}
\resizebox{\linewidth}{!}{
\begin{tabular}{c|c|cccccc}
\toprule
\textbf{Datasets} & \textbf{Feature} & \textbf{TGAT} & \textbf{CAWN} & \textbf{TCL} & \textbf{GraphMixer} & \textbf{DyGFormer} & \textbf{TST-MF} \\
\midrule
\multirow{4}{*}{GDELT} & Zero Filling & 50.86 $\pm$ 0.21 & \textbf{53.40 $\pm$ 0.11} & 49.77 $\pm$ 0.00 & 52.62 $\pm$ 0.66 & 52.85 $\pm$ 0.22 & \underline{53.27 $\pm$ 0.34} \\
 & Mean Filling & 50.52 $\pm$ 0.59 & \textbf{53.38 $\pm$ 0.09} & 49.80 $\pm$ 0.05 & 52.24 $\pm$ 0.30 & 52.87 $\pm$ 0.03 & \underline{53.22 $\pm$ 0.74} \\
 & Learned Embedding & 50.69 $\pm$ 0.08 & \textbf{53.26 $\pm$ 0.16} & 49.79 $\pm$ 0.04 & 52.95 $\pm$ 0.19 & 52.75 $\pm$ 0.14 & \underline{53.09 $\pm$ 0.55} \\
 & Duplication & 50.81 $\pm$ 0.16 & \underline{53.45 $\pm$ 0.10} & 49.77 $\pm$ 0.00 & 52.66 $\pm$ 0.86 & 52.81 $\pm$ 0.07 & \textbf{53.64 $\pm$ 0.03} \\
\midrule
\multirow{4}{*}{GMCT} & Zero Filling & 59.53 $\pm$ 0.58 & 59.19 $\pm$ 0.09 & 59.05 $\pm$ 0.64 & \textbf{60.02 $\pm$ 0.33} & \underline{59.75 $\pm$ 0.10} & 55.43 $\pm$ 1.24 \\
 & Mean Filling & 59.25 $\pm$ 0.50 & 59.42 $\pm$ 0.07 & 58.13 $\pm$ 0.49 & \textbf{59.78 $\pm$ 0.12} & \underline{59.60 $\pm$ 0.26} & 55.67 $\pm$ 0.54 \\
 & Learned Embedding & 58.93 $\pm$ 0.57 & 59.18 $\pm$ 0.06 & 58.50 $\pm$ 0.36 & \textbf{59.84 $\pm$ 0.18} & \underline{59.78 $\pm$ 0.21} & 55.55 $\pm$ 1.37 \\
\bottomrule
\end{tabular}
}
\end{table*}

\begin{table*}[t]
\centering
\small
\caption{Hits@1 (\%) for node retrieval under different missing-modality filling strategies. ``\textit{tr.}'' and ``\textit{in.}'' refer to the transductive and inductive settings, respectively. The best results are highlighted in bold and the second-best results are underlined.}
\label{tab:supp_missing_nr_h1}
\resizebox{\linewidth}{!}{
\begin{tabular}{c|c|c|cccccc}
\toprule
 & \textbf{Datasets} & \textbf{Feature} & \textbf{TGAT} & \textbf{CAWN} & \textbf{TCL} & \textbf{GraphMixer} & \textbf{DyGFormer} & \textbf{TST-MF} \\
\midrule
\multirow{20}{*}{\textit{tr.}} & \multirow{4}{*}{GDELT} & Zero Filling & 31.11 $\pm$ 1.23 & \underline{33.56 $\pm$ 0.81} & 31.62 $\pm$ 0.04 & 30.92 $\pm$ 0.34 & 30.98 $\pm$ 3.35 & \textbf{38.32 $\pm$ 0.36} \\
 &  & Mean Filling & 29.89 $\pm$ 0.96 & \underline{33.26 $\pm$ 0.78} & 31.71 $\pm$ 0.00 & 30.97 $\pm$ 0.07 & 31.05 $\pm$ 1.39 & \textbf{38.36 $\pm$ 0.16} \\
 &  & Learned Embedding & 30.05 $\pm$ 0.48 & 33.46 $\pm$ 1.22 & 31.61 $\pm$ 0.01 & 31.02 $\pm$ 0.25 & \underline{33.92 $\pm$ 3.22} & \textbf{38.09 $\pm$ 1.07} \\
 &  & Duplication & 30.42 $\pm$ 1.64 & 33.65 $\pm$ 1.09 & 31.53 $\pm$ 0.31 & 31.11 $\pm$ 0.64 & \underline{34.19 $\pm$ 4.03} & \textbf{38.31 $\pm$ 0.74} \\
\cmidrule{2-9}
 & \multirow{4}{*}{GMCT} & Zero Filling & 14.73 $\pm$ 4.35 & 11.54 $\pm$ 0.16 & 11.29 $\pm$ 0.25 & 11.62 $\pm$ 0.25 & \textbf{91.12 $\pm$ 7.18} & \underline{53.84 $\pm$ 11.14} \\
 &  & Mean Filling & 17.04 $\pm$ 1.98 & 11.58 $\pm$ 0.16 & 11.36 $\pm$ 0.23 & 11.53 $\pm$ 0.07 & \textbf{93.53 $\pm$ 1.50} & \underline{65.77 $\pm$ 2.01} \\
 &  & Learned Embedding & 13.80 $\pm$ 3.21 & 11.07 $\pm$ 0.27 & 11.57 $\pm$ 0.33 & 10.50 $\pm$ 2.46 & \textbf{91.94 $\pm$ 1.93} & \underline{60.24 $\pm$ 14.12} \\
\bottomrule
\end{tabular}
}
\end{table*}

\begin{table*}[t]
\centering
\small
\caption{Hits@3 (\%) for node retrieval under different missing-modality filling strategies. ``\textit{tr.}'' and ``\textit{in.}'' refer to the transductive and inductive settings, respectively. The best results are highlighted in bold and the second-best results are underlined.}
\label{tab:supp_missing_nr_h3}
\resizebox{\linewidth}{!}{
\begin{tabular}{c|c|c|cccccc}
\toprule
 & \textbf{Datasets} & \textbf{Feature} & \textbf{TGAT} & \textbf{CAWN} & \textbf{TCL} & \textbf{GraphMixer} & \textbf{DyGFormer} & \textbf{TST-MF} \\
\midrule
\multirow{20}{*}{\textit{tr.}} & \multirow{4}{*}{GDELT} & Zero Filling & 52.29 $\pm$ 0.85 & \underline{56.05 $\pm$ 0.66} & 53.79 $\pm$ 0.13 & 51.47 $\pm$ 1.10 & 45.49 $\pm$ 7.59 & \textbf{59.28 $\pm$ 0.39} \\
 &  & Mean Filling & 50.96 $\pm$ 1.07 & \underline{55.35 $\pm$ 0.45} & 53.63 $\pm$ 0.38 & 51.93 $\pm$ 0.55 & 46.53 $\pm$ 3.51 & \textbf{59.50 $\pm$ 0.02} \\
 &  & Learned Embedding & 50.86 $\pm$ 1.06 & \underline{55.71 $\pm$ 1.37} & 53.61 $\pm$ 0.16 & 52.29 $\pm$ 0.38 & 51.30 $\pm$ 6.33 & \textbf{59.07 $\pm$ 1.45} \\
 &  & Duplication & 51.56 $\pm$ 2.13 & \underline{55.65 $\pm$ 1.06} & 53.46 $\pm$ 0.21 & 51.71 $\pm$ 2.01 & 52.06 $\pm$ 6.41 & \textbf{59.05 $\pm$ 0.36} \\
\cmidrule{2-9}
 & \multirow{4}{*}{GMCT} & Zero Filling & 30.79 $\pm$ 6.07 & 25.57 $\pm$ 0.12 & 25.54 $\pm$ 0.38 & 25.81 $\pm$ 0.74 & \textbf{99.98 $\pm$ 0.03} & \underline{97.40 $\pm$ 1.97} \\
 &  & Mean Filling & 33.83 $\pm$ 3.37 & 25.51 $\pm$ 0.14 & 24.72 $\pm$ 0.23 & 26.26 $\pm$ 0.98 & \textbf{100.00 $\pm$ 0.00} & \underline{99.31 $\pm$ 0.08} \\
 &  & Learned Embedding & 29.24 $\pm$ 4.29 & 24.97 $\pm$ 0.06 & 25.25 $\pm$ 0.31 & 24.20 $\pm$ 2.36 & \textbf{100.00 $\pm$ 0.00} & \underline{97.12 $\pm$ 4.11} \\
\bottomrule
\end{tabular}
}
\end{table*}

\begin{table*}[t]
\centering
\small
\caption{Hits@10 (\%) for node retrieval under different missing-modality filling strategies. ``\textit{tr.}'' and ``\textit{in.}'' refer to the transductive and inductive settings, respectively. The best results are highlighted in bold and the second-best results are underlined.}
\label{tab:supp_missing_nr_h10}
\resizebox{\linewidth}{!}{
\begin{tabular}{c|c|c|cccccc}
\toprule
 & \textbf{Datasets} & \textbf{Feature} & \textbf{TGAT} & \textbf{CAWN} & \textbf{TCL} & \textbf{GraphMixer} & \textbf{DyGFormer} & \textbf{TST-MF} \\
\midrule
\multirow{20}{*}{\textit{tr.}} & \multirow{4}{*}{GDELT} & Zero Filling & 70.55 $\pm$ 2.09 & \underline{79.20 $\pm$ 0.13} & 75.11 $\pm$ 0.31 & 70.79 $\pm$ 1.22 & 63.87 $\pm$ 10.27 & \textbf{81.83 $\pm$ 0.38} \\
 &  & Mean Filling & 70.47 $\pm$ 0.99 & \underline{79.05 $\pm$ 0.27} & 74.92 $\pm$ 0.59 & 71.28 $\pm$ 1.57 & 64.12 $\pm$ 5.79 & \textbf{82.01 $\pm$ 0.66} \\
 &  & Learned Embedding & 69.57 $\pm$ 2.06 & \underline{79.41 $\pm$ 0.70} & 74.84 $\pm$ 0.22 & 71.74 $\pm$ 1.53 & 70.45 $\pm$ 8.88 & \textbf{81.67 $\pm$ 0.72} \\
 &  & Duplication & 70.79 $\pm$ 1.04 & \underline{79.08 $\pm$ 0.35} & 74.66 $\pm$ 0.43 & 72.41 $\pm$ 1.16 & 71.34 $\pm$ 7.82 & \textbf{81.62 $\pm$ 0.33} \\
\cmidrule{2-9}
 & \multirow{4}{*}{GMCT} & Zero Filling & \underline{54.67 $\pm$ 4.52} & 50.77 $\pm$ 0.04 & 50.20 $\pm$ 0.20 & 50.84 $\pm$ 0.43 & \textbf{100.00 $\pm$ 0.00} & \textbf{100.00 $\pm$ 0.00} \\
 &  & Mean Filling & \underline{57.29 $\pm$ 3.80} & 50.72 $\pm$ 0.11 & 49.83 $\pm$ 0.45 & 50.91 $\pm$ 0.33 & \textbf{100.00 $\pm$ 0.00} & \textbf{100.00 $\pm$ 0.00} \\
 &  & Learned Embedding & 52.68 $\pm$ 2.56 & 50.53 $\pm$ 0.08 & 49.78 $\pm$ 0.41 & 50.34 $\pm$ 0.35 & \textbf{100.00 $\pm$ 0.00} & \underline{99.97 $\pm$ 0.04} \\
\bottomrule
\end{tabular}
}
\end{table*}

\begin{table*}[t]
\centering
\small
\caption{Hits@10 (\%) for recurrence-aware node retrieval with historical negative sampling. ``\textit{tr.}'' and ``\textit{in.}'' refer to the transductive and inductive settings, respectively. The best results are highlighted in bold and the second-best results are underlined.}
\label{tab:supp_recurrence_h10}
\resizebox{\linewidth}{!}{
\begin{tabular}{c|c|c|cccccc}
\toprule
 & \textbf{Datasets} & \textbf{Feature} & \textbf{TGAT} & \textbf{CAWN} & \textbf{TCL} & \textbf{GraphMixer} & \textbf{DyGFormer} & \textbf{TST-MF} \\
\midrule
\multirow{15}{*}{\textit{tr.}} & \multirow{3}{*}{GDELT} & Random & 20.78 $\pm$ 0.26 & 26.18 $\pm$ 0.80 & 20.98 $\pm$ 0.16 & 20.70 $\pm$ 0.25 & \textbf{33.32 $\pm$ 0.57} & \underline{27.27 $\pm$ 0.78} \\
 &  & Text & 20.77 $\pm$ 1.49 & 27.81 $\pm$ 1.13 & 20.65 $\pm$ 0.09 & 21.58 $\pm$ 1.52 & \textbf{35.39 $\pm$ 0.49} & \underline{34.81 $\pm$ 1.46} \\
 &  & Text+Image & 22.70 $\pm$ 1.50 & 27.73 $\pm$ 0.91 & 20.75 $\pm$ 0.11 & 20.82 $\pm$ 0.25 & \underline{35.56 $\pm$ 0.34} & \textbf{36.03 $\pm$ 0.78} \\
\cmidrule{2-9}
 & \multirow{3}{*}{GMCT} & Random & 41.08 $\pm$ 3.52 & 44.08 $\pm$ 0.09 & 44.04 $\pm$ 0.14 & 44.06 $\pm$ 0.05 & \textbf{99.91 $\pm$ 0.12} & \underline{47.54 $\pm$ 30.15} \\
 &  & Text & 47.63 $\pm$ 4.12 & 44.15 $\pm$ 0.03 & 43.91 $\pm$ 0.22 & 25.25 $\pm$ 16.46 & \textbf{99.96 $\pm$ 0.06} & \underline{95.85 $\pm$ 3.10} \\
 &  & Text+Image & 45.95 $\pm$ 1.68 & 44.19 $\pm$ 0.06 & 43.99 $\pm$ 0.18 & 34.15 $\pm$ 11.72 & \textbf{99.99 $\pm$ 0.02} & \underline{81.52 $\pm$ 3.04} \\
\cmidrule{2-9}
 & \multirow{3}{*}{Amazon - Industrial} & Random & \textbf{45.17 $\pm$ 3.35} & \underline{41.86 $\pm$ 1.79} & 23.97 $\pm$ 5.51 & 34.07 $\pm$ 2.97 & 23.54 $\pm$ 4.73 & 31.44 $\pm$ 0.75 \\
\bottomrule
\end{tabular}
}
\end{table*}

\begin{table*}[t]
\centering
\small
\caption{Repeated Hits@10 (\%) for recurrence-aware node retrieval with historical negative sampling. ``\textit{tr.}'' and ``\textit{in.}'' refer to the transductive and inductive settings, respectively. The best results are highlighted in bold and the second-best results are underlined.}
\label{tab:supp_recurrence_repeated_h10}
\resizebox{\linewidth}{!}{
\begin{tabular}{c|c|c|cccccc}
\toprule
 & \textbf{Datasets} & \textbf{Feature} & \textbf{TGAT} & \textbf{CAWN} & \textbf{TCL} & \textbf{GraphMixer} & \textbf{DyGFormer} & \textbf{TST-MF} \\
\midrule
\multirow{15}{*}{\textit{tr.}} & \multirow{3}{*}{GDELT} & Random & 27.26 $\pm$ 0.32 & 35.93 $\pm$ 1.14 & 27.38 $\pm$ 0.22 & 27.11 $\pm$ 0.33 & \textbf{45.43 $\pm$ 0.76} & \underline{37.46 $\pm$ 1.11} \\
 &  & Text & 27.07 $\pm$ 2.30 & 38.30 $\pm$ 1.62 & 26.88 $\pm$ 0.12 & 28.35 $\pm$ 2.12 & \textbf{47.80 $\pm$ 0.67} & \underline{46.73 $\pm$ 1.94} \\
 &  & Text+Image & 29.62 $\pm$ 1.98 & 38.03 $\pm$ 1.32 & 27.06 $\pm$ 0.12 & 27.36 $\pm$ 0.32 & \textbf{48.13 $\pm$ 0.46} & \underline{48.08 $\pm$ 0.95} \\
\cmidrule{2-9}
 & \multirow{3}{*}{GMCT} & Random & 33.76 $\pm$ 2.96 & 37.38 $\pm$ 0.28 & 38.49 $\pm$ 0.04 & 38.27 $\pm$ 0.20 & \textbf{99.03 $\pm$ 0.10} & \underline{45.75 $\pm$ 28.91} \\
 &  & Text & 42.04 $\pm$ 4.45 & 37.58 $\pm$ 0.20 & 37.74 $\pm$ 0.32 & 24.11 $\pm$ 12.57 & \textbf{99.05 $\pm$ 0.06} & \underline{94.89 $\pm$ 3.06} \\
 &  & Text+Image & 40.61 $\pm$ 2.61 & 37.54 $\pm$ 0.22 & 38.32 $\pm$ 0.13 & 30.79 $\pm$ 8.96 & \textbf{99.09 $\pm$ 0.00} & \underline{81.00 $\pm$ 2.87} \\
\cmidrule{2-9}
 & \multirow{3}{*}{Amazon - Industrial} & Random & \textbf{60.92 $\pm$ 3.03} & \underline{58.68 $\pm$ 1.15} & 38.04 $\pm$ 14.49 & 55.23 $\pm$ 1.60 & 42.04 $\pm$ 4.60 & 51.49 $\pm$ 0.96 \\
\bottomrule
\end{tabular}
}
\end{table*}

\begin{table*}[t]
\centering
\small
\caption{Novel Hits@10 (\%) for recurrence-aware node retrieval with historical negative sampling. ``\textit{tr.}'' and ``\textit{in.}'' refer to the transductive and inductive settings, respectively. The best results are highlighted in bold and the second-best results are underlined.}
\label{tab:supp_recurrence_novel_h10}
\resizebox{\linewidth}{!}{
\begin{tabular}{c|c|c|cccccc}
\toprule
 & \textbf{Datasets} & \textbf{Feature} & \textbf{TGAT} & \textbf{CAWN} & \textbf{TCL} & \textbf{GraphMixer} & \textbf{DyGFormer} & \textbf{TST-MF} \\
\midrule
\multirow{15}{*}{\textit{tr.}} & \multirow{3}{*}{GDELT} & Random & 8.77 $\pm$ 0.16 & 7.83 $\pm$ 0.20 & \underline{9.04 $\pm$ 0.08} & 8.81 $\pm$ 0.11 & \textbf{10.42 $\pm$ 0.23} & 8.27 $\pm$ 0.15 \\
 &  & Text & 9.09 $\pm$ 0.04 & 8.06 $\pm$ 0.28 & 9.04 $\pm$ 0.03 & 8.98 $\pm$ 0.36 & \underline{11.96 $\pm$ 0.55} & \textbf{12.44 $\pm$ 0.55} \\
 &  & Text+Image & 9.79 $\pm$ 0.66 & 8.34 $\pm$ 0.64 & 9.03 $\pm$ 0.07 & 8.69 $\pm$ 0.21 & \underline{11.81 $\pm$ 0.26} & \textbf{13.40 $\pm$ 0.60} \\
\cmidrule{2-9}
 & \multirow{3}{*}{GMCT} & Random & 41.22 $\pm$ 3.52 & 44.20 $\pm$ 0.09 & 44.15 $\pm$ 0.15 & 44.17 $\pm$ 0.04 & \textbf{99.91 $\pm$ 0.12} & \underline{47.41 $\pm$ 30.28} \\
 &  & Text & 47.73 $\pm$ 4.12 & 44.27 $\pm$ 0.02 & 44.03 $\pm$ 0.23 & 25.27 $\pm$ 16.54 & \textbf{99.96 $\pm$ 0.06} & \underline{95.89 $\pm$ 3.10} \\
 &  & Text+Image & 46.03 $\pm$ 1.66 & 44.30 $\pm$ 0.06 & 44.10 $\pm$ 0.17 & 34.22 $\pm$ 11.77 & \textbf{99.99 $\pm$ 0.02} & \underline{81.45 $\pm$ 3.09} \\
\cmidrule{2-9}
 & \multirow{3}{*}{Amazon - Industrial} & Random & \textbf{44.79 $\pm$ 3.36} & \underline{41.48 $\pm$ 1.81} & 23.67 $\pm$ 5.40 & 33.60 $\pm$ 3.00 & 23.14 $\pm$ 4.72 & 30.84 $\pm$ 0.78 \\
\bottomrule
\end{tabular}
}
\end{table*}

\begin{table*}[h]
\centering
\small
\caption{Instance-level utility of each modality for TST-MF under the Text+Image setting. For a test instance $i$ and modality $m$, $\Delta_{i,m}=\ell(f_{\setminus m}(i),y_i)-\ell(f(i),y_i)$, where $f_{\setminus m}$ zeroes modality $m$ in every pathway of the model and $\ell$ is the negative log-likelihood of the true interaction (link prediction) or of the true class (edge classification); $\Delta_{i,m}>0$ means that the modality helps the prediction.}
\label{tab:modality_utility}
\resizebox{\linewidth}{!}{
\begin{tabular}{c|c|cc|cc}
\toprule
 & & \multicolumn{2}{c|}{\textbf{Mean $\Delta$ [95\% CI]}} & \multicolumn{2}{c|}{\textbf{Share of $\Delta>0$ (\%)}} \\
\textbf{Datasets} & \textbf{Task} & Text & Image & Text & Image \\
\midrule
\multirow{3}{*}{GDELT} & LP (\textit{tr.}) & -0.096 {\scriptsize [-0.102, -0.092]} & -0.103 {\scriptsize [-0.130, -0.084]} & 42.1 & 36.6 \\
 & LP (\textit{in.}) & -0.124 {\scriptsize [-0.130, -0.117]} & -0.131 {\scriptsize [-0.158, -0.108]} & 37.7 & 33.3\\
 & EC & -0.004 {\scriptsize [-0.014, +0.003]} & +0.000 {\scriptsize [-0.006, +0.006]} & 46.8 & 48.1 \\
\midrule
\multirow{3}{*}{GMCT} & LP (\textit{tr.}) & +0.149 {\scriptsize [+0.003, +0.406]} & +0.282 {\scriptsize [+0.006, +0.818]} & 76.6 & 77.2 \\
 & LP (\textit{in.}) & +0.163 {\scriptsize [+0.003, +0.443]} & +0.308 {\scriptsize [+0.006, +0.889]} & 78.6 & 79.3 \\
 & EC & +0.011 {\scriptsize [+0.010, +0.012]} & +0.008 {\scriptsize [+0.006, +0.010]} & 44.1 & 56.8 \\
\bottomrule
\end{tabular}}
\end{table*}

\clearpage
\section{Ablation Study}
\label{app:ablation}
We conduct ablation studies on our proposed \textbf{TST-MF} method to verify the effectiveness of each design choice. The result is given in Table \ref{tab:abl} and \ref{tab:modality_utility}.

\begin{table*}[h] 
    \centering
    \caption{Ablation Study over Link Prediction Task. Four variants are experimented.  “w/o target” removes target guidance from the context cross-attention layer; “w/o hier.” replaces hierarchical fusion with a single-stage alternative; “w/o spatial” removes the structural-context branch and “w/o historical” removes the interaction-history branch. Best results are \textbf{bolded}.} 
    \label{tab:abl}

    \resizebox{\linewidth}{!}{ 
    
    \begin{tabular}{c|c|c|ccccc} \toprule  
    & \textbf{Datasets} & \textbf{Metric} & \textbf{TST-MF} & \textbf{w/o target} & \textbf{w/o hier.} & \textbf{w/o spatial} & \textbf{w/o historical} \\ 
    \midrule

    \multirow{5}{*}{\textit{tr.}} 
    & \multirow{2}{*}{GDELT} 
    & Precision & \textbf{93.98 $\pm$ 0.05} & 92.08 $\pm$ 0.21 & 91.90 $\pm$ 0.15 & 92.06 $\pm$ 0.22 & 93.95 $\pm$ 0.06 \\ 
    & & AUC-ROC & \textbf{93.71 $\pm$ 0.08} & 92.78 $\pm$ 0.18 & 91.54 $\pm$ 0.20 & 92.74 $\pm$ 0.29 & 91.78 $\pm$ 0.09 \\
    \cmidrule{2-8}

    & \multirow{2}{*}{GMCT} 
    & Precision & \textbf{89.08 $\pm$ 0.72} & 88.83 $\pm$ 1.24 & 82.28 $\pm$ 3.99 & 88.46 $\pm$ 0.66 & 89.03 $\pm$ 0.34\\ 
    & & AUC-ROC & \textbf{92.49 $\pm$ 0.66} & 91.67 $\pm$ 0.89 & 82.28 $\pm$ 4.06 & 91.33 $\pm$ 0.88 & 91.87 $\pm$ 0.04 \\
    \cmidrule{2-8}

    & \multirow{2}{*}{Amazon - Industrial} 
    & Precision & \textbf{94.10 $\pm$ 0.12} & 94.01 $\pm$ 0.28 & 93.56 $\pm$ 0.50 & 94.01 $\pm$ 0.23 & 93.99 $\pm$ 0.05 \\ 
    & & AUC-ROC & \textbf{94.26 $\pm$ 0.16} & 93.23 $\pm$ 0.27 & 93.94 $\pm$ 0.22 &93.25 $\pm$ 0.19 & 93.38 $\pm$ 0.11 \\
    \cmidrule{2-8}

    & \multirow{2}{*}{Amazon - Office} 
    & Precision & \textbf{92.67 $\pm$ 0.10} & 91.86 $\pm$ 0.11 & 91.94 $\pm$ 0.22 & 92.01 $\pm$ 0.29 & 91.40 $\pm$ 0.38 \\ 
    & & AUC-ROC & \textbf{93.23 $\pm$ 0.43} & 92.97 $\pm$ 0.35 & 92.72 $\pm$ 0.15 & 92.87 $\pm$ 0.15 & 92.62 $\pm$ 0.21 \\
    \cmidrule{2-8}

    & \multirow{2}{*}{Yelp} 
    & Precision & \textbf{92.65 $\pm$ 0.04} & 92.51 $\pm$ 0.41 & 92.31 $\pm$ 0.07 & 92.32 $\pm$ 0.09 & 92.28 $\pm$ 0.10 \\ 
    & & AUC-ROC & \textbf{93.83 $\pm$ 0.02} & 92.86 $\pm$ 0.26 & 92.64 $\pm$ 0.07 & 92.69 $\pm$ 0.10 & 92.61 $\pm$ 0.11 \\

    \cmidrule{1-8}

    \multirow{5}{*}{\textit{in.}} 
    & \multirow{2}{*}{GDELT} 
    & Precision & 85.44 $\pm$ 0.14 & 84.69 $\pm$ 0.52 & \textbf{85.50 $\pm$ 0.27} & 84.44 $\pm$ 0.22 & 84.32 $\pm$ 0.19 \\ 
    & & AUC-ROC & \textbf{84.51 $\pm$ 0.11} & 84.46 $\pm$ 0.48 & 84.48 $\pm$ 0.34 & 84.42 $\pm$ 0.13 & 84.41 $\pm$ 0.06 \\
    \cmidrule{2-8}

    & \multirow{2}{*}{GMCT} 
    & Precision & \textbf{65.49 $\pm$ 1.00} & 64.92 $\pm$ 2.72 & 64.80 $\pm$ 6.82 & 64.57 $\pm$ 1.03 & 64.78 $\pm$ 0.44 \\ 
    & & AUC-ROC & \textbf{67.77 $\pm$ 1.48} & 66.30 $\pm$ 2.30 & 67.30 $\pm$ 6.50 & 66.62 $\pm$ 1.86 & 67.56 $\pm$ 0.19 \\
    \cmidrule{2-8}

    & \multirow{2}{*}{Amazon - Industrial} 
    & Precision & \textbf{81.95 $\pm$ 0.54} & 81.74 $\pm$ 0.93 & 80.72 $\pm$ 1.98 & 81.70 $\pm$ 0.89 & 81.34 $\pm$ 0.48 \\ 
    & & AUC-ROC & \textbf{80.00 $\pm$ 0.60} & 79.64 $\pm$ 1.26 & 79.03 $\pm$ 1.08 & 79.52 $\pm$ 0.99 & 79.10 $\pm$ 0.17 \\
    \cmidrule{2-8}

    & \multirow{2}{*}{Amazon - Office} 
    & Precision & \textbf{78.16 $\pm$ 0.31} & 76.88 $\pm$ 1.49 & 77.27 $\pm$ 0.71 & 77.11 $\pm$ 1.00 & 75.05 $\pm$ 1.03\\ 
    & & AUC-ROC & \textbf{75.38 $\pm$ 1.60} & 74.98 $\pm$ 1.33 & 74.59 $\pm$ 0.45 & 74.73 $\pm$ 0.78 & 73.28 $\pm$ 1.05 \\
    \cmidrule{2-8}

    & \multirow{2}{*}{Yelp} 
    & Precision & \textbf{89.39 $\pm$ 0.05} & 89.24 $\pm$ 0.37 & 88.99 $\pm$ 0.11 & 88.98 $\pm$ 0.10 & 88.92 $\pm$ 0.01\\ 
    & & AUC-ROC & \textbf{89.10 $\pm$ 0.05} & 89.28 $\pm$ 0.28 & 88.92 $\pm$ 0.12 & 89.01 $\pm$ 0.10 & 88.89 $\pm$ 0.09 \\
    
    \bottomrule
    \end{tabular}}
\end{table*}

\paragraph{Effect of target guidance.} When we replace the context cross-attention layer with a direct self-attention layer for the concatenated historical and structural embeddings, performance drops consistently across datasets and tasks. This proves that multimodal utility in dynamic graphs is target-dependent. The same modality can be helpful for some prediction targets, but distracting for others. Without target-guided querying, the fusion module tends to produce a target-agnostic mixture, which amplifies irrelevant signals and weakens the model’s ability to focus on evidence that is most informative for the queried target.

\paragraph{Effect of the hierarchical architecture.} Replacing the hierarchical two-stage fusion with a single-stage attention (querying the concatenated structural, historical, and multimodal features with the prediction targets) consistently degrades performance. This suggests that one-shot fusion is more susceptible to modality noise and misalignment, whereas the two-stage design first forms a target-aware spatiotemporal context and then uses it to regulate cross-modal interplay, yielding more stable fusion under dynamic graphs.

\paragraph{Effect of spatial vs. historical context.} Finally, removing either the spatial neighborhood context or the historical temporal context harms performance. These results confirm that multimodal fusion in dynamic graphs should be conditioned on both \emph{where} the target lies in the evolving graph and \emph{when} it is observed along its interaction history. Relying on only one source of context provides incomplete cues for deciding which modality to trust.

\paragraph{Effect of modality.} The effect of a modality is neither consistent in direction nor uniform across instances. For example, for link prediction, the same modalities are beneficial on average on Yelp but harmful on average on GDELT, and even where they help on average, they degrade a substantial share of individual predictions. A fusion rule that treats each modality as uniformly informative is therefore ill-suited to multimodal dynamic graphs, which motivates conditioning the fusion on the prediction target, as in \textbf{TST-MF}.

\clearpage

\section{Broader Impacts}
\label{app:br}
This work aims to support research on multimodal dynamic graph learning by providing datasets, tasks, baselines, and a target-aware fusion framework. MM-DyGraph can help researchers evaluate when multimodal information is useful in dynamic relational systems, rather than assuming that adding more modalities always improves performance. By revealing target-level inconsistency, it encourages more careful model design and evaluation for applications such as recommendation, urban computing, and event analysis.

However, multimodal dynamic graph learning may also raise societal risks if used without safeguards. Dynamic graphs can encode evolving relationships among users, organizations, locations, and events, while text and image attributes may make learned representations more semantically informative. As a result, models trained on such data could potentially amplify biases, support user profiling, or produce misleading predictions when relying on noisy or spurious multimodal cues.

To mitigate these risks, MM-DyGraph is intended for research evaluation rather than direct high-stakes deployment. Users should follow data licenses and terms of use, avoid attempts to identify individuals, and refrain from using the benchmark for surveillance or automated profiling.

\section{Limitations}
\label{app:limitations}
Although MM-DyGraph provides a benchmark for multimodal dynamic graph learning, several limitations remain. First, dataset-specific pruning and VLM-based filtering improve multimodal quality but may bias the data toward clearer, higher-quality reviews or interactions, underrepresenting noisy, sparse, or weakly aligned real-world samples. Second, MM-DyGraph mainly focuses on text and image modalities, leaving other signals such as audio, video, tabular metadata, and sensor data unexplored. Third, this work focuses on dynamic graph structure with multimodal attributes and does not systematically cover successive updates to the raw text or images of the same entity. Datasets with explicit entity-level feature updates are an important direction for future works. Finally, VLM-based filtering is model- and prompt-dependent and cannot guarantee perfect cross-modal alignment. 


\end{document}